\documentclass[10pt]{article}

\usepackage[numbers]{natbib}
\usepackage{amsmath}
\usepackage{amssymb}

\usepackage{bm}
\usepackage{algorithm}
\usepackage[noend]{algpseudocode}
\usepackage{graphicx} 
\usepackage{subcaption} 
\usepackage{sidecap}
\usepackage{multirow}
\usepackage{hyperref}

\usepackage{lineno}

\usepackage{microtype}
\DisableLigatures[f]{encoding = *, family = * }

\usepackage[labelfont=bf,labelsep=period,justification=raggedright]{caption}

\makeatletter
\renewcommand{\@biblabel}[1]{\quad#1.}
\makeatother

\date{}

\newcommand{\lp}{\left(}
\newcommand{\rp}{\right)}
\newcommand{\lb}{\left[}
\newcommand{\rb}{\right]}

\newcommand{\eps}{\epsilon}

\newcommand{\y}{{\bf y}}
\newcommand{\ys}{\y^{(s)}}
\newcommand{\yp}{\y^{\prime}}
\newcommand{\yps}{\y^{\prime(s)}}
\newcommand{\ypS}{\y^{\prime(S)}}

\newcommand{\yone}{\y^{(1)}}
\newcommand{\yS}{\y^{(S)}}
\newcommand{\ypone}{\y^{'(1)}}

\newcommand{\thetap}{\theta^{\prime}}

\newcommand{\thetav}{{\bm \theta}}
\newcommand{\thetapv}{{\bm \theta^{\prime}}}
\newcommand{\thetavp}{\thetapv}
\newcommand{\muhattheta}{\hat{{\bm \mu}}_{\thetav}}

\newcommand{\muhatthetaj}{\hat{\mu}_{j}}
\newcommand{\sigmahatthetaj}{\hat{\sigma}^2_{j}}

\newcommand{\yj}{y_j}

\newcommand{\ymu}{\muhattheta}
\newcommand{\simsim}{\overset{\simulator}{\sim}}
\newcommand{\ystar}{\y^{\star}}
\newcommand{\thetastar}{\thetav^{\star}}

\newcommand{\ystarj}{\yj^{\star}}

\newcommand{\ymc}{\y^{(s)}}
\newcommand{\epsvec}{ {\bm \eps}}
\newcommand{\epsj}{\eps_j}
\newcommand{\F}{F}

\DeclareMathOperator\simulator{sim}

\DeclareMathOperator\mean{mean}
\DeclareMathOperator\median{median}
\DeclareMathOperator\mode{mode}

\newcommand{\E}{E}
\newcommand{\yemaj}{\yj^{\textsc{ema}}}
\newcommand{\yema}{\y^{\textsc{ema}}}
\newcommand{\epsemaj}{\epsj^{\textsc{ema}}}
\newcommand{\epsema}{\epsvec^{\textsc{ema}}}
\newcommand{\epsmin}{\epsvec^{\textsc{min}}}
\newcommand{\epsminj}{\epsj^{\textsc{min}}}
\newcommand{\epsquantile}{\beta}
\newcommand{\geom}{{\bf g}}

\title{POPE: Post Optimization Posterior Evaluation \\ of Likelihood Free Models}

\author{
Edward Meeds \\
Informatics Institute \\
University of Amsterdam\\
Amsterdam, The Netherlands \\
\texttt{tmeeds@gmail.com} \\
\And
Max Welling \\
Informatics Institute \\
University of Amsterdam\\
Amsterdam, The Netherlands \\
\texttt{welling.max@gmail.com}
}

\begin{document}

\begin{flushleft}
{\Large
\textbf{POPE: Post Optimization Posterior Evaluation \\ of Likelihood Free Models}
}
\\
Edward Meeds$^{1\ast}$, 
Michael Chiang$^{2}$, 
Mary Lee$^{3}$, 
Olivier Cinquin$^{2}$, 
John Lowengrub$^{3}$, 
Max Welling$^{1,4}$
\\
\bf{1} Informatics Institute, University of Amsterdam, Amsterdam, The Netherlands
\\
\bf{2} School of Biological Sciences, University of California, Irvine, CA, USA
\\
\bf{3} Department of Mathematics, University of California, Irvine, CA, USA
\\
\bf{4} Donald Bren School of Informatics, University of California, Irvine, CA, USA
\\
$\ast$ E-mail: Corresponding E.W.F.Meeds@uva.nl
\end{flushleft}

\section*{Abstract}
In many domains, scientists build complex simulators of natural phenomena that encode their hypotheses about the underlying processes. These simulators can be deterministic or stochastic, fast or slow, constrained or unconstrained, and so on. Optimizing the simulators with respect to a set of parameter values is common practice, resulting in a single parameter setting that minimizes an objective subject to constraints. We propose a post optimization posterior analysis that computes and visualizes all the models that can generate equally good or better simulation results, subject to constraints. These {\em optimization posteriors} are desirable for a number of reasons among which easy interpretability, automatic parameter sensitivity and correlation analysis and posterior predictive analysis.  We develop a new sampling framework based on approximate Bayesian computation (ABC) with one-sided kernels. In collaboration with two groups of scientists we applied POPE to two important biological simulators: a fast and stochastic simulator of stem-cell cycling and a slow and deterministic simulator of tumor growth patterns.


\section*{Introduction}
In science and industry alike, modelers express their expert knowledge by building a simulator of the phenomenon of interest. There is an enormous variety of such simulators, deterministic or stochastic, fast or slow, with or without constraints. For most simulators, e.g. driven by stochastic partial differential equations, it is impossible to write down an expression for the likelihood, which can make it highly challenging to optimize the simulator over its free parameters. This ``blind optimization problem" is receiving increasing attention in the machine learning community \cite{lizotte2008,osborne2009,snoek:2012}.

However, even if the optimal parameter value $\thetastar$ is found, this leaves the scientist still in the dark with respect to important questions such as: ``Which parameters are correlated?"; ``Which parameters are robust and which are sensitive?"; ``Is my model overfitting, underfitting or just right"? We believe that methods capable of handling these type of questions post optimization are essential to the field of simulation-based modeling. In this paper we propose a new Bayesian framework that allows the scientist to answer these questions by approximating through sampling the posterior distribution of all parameters that may result in equally good or better models.  This ``Post Optimization Posterior Evaluation" (POPE) is different from standard ABC \cite{marjoram2003markov,Wilkinson2013,sisson:2010} in that standard ABC compares simulator outcomes with observations while POPE reasons about an optimization problem (subject to constraints) without the need for observations. While different philosophically, POPE can be implemented by using one-sided kernels within ABC.

POPE was developed in close collaboration with a number of scientists, and has a number of properties that are beneficial to their work: 1) the posterior distribution over parameters has a clear and interpretable meaning and can be used to suggest alternative parameters to explore, 2) POPE can handle multiple objectives and constraints, 3) unlike most standard optimization methods, POPE can handle simulators with stochastic outputs and complicated input or outputs constraints, 4) POPE can handle multimodal posterior distributions, 5) as part of its computation POPE will generate posterior predictive samples that can be used to evaluate the model fit, and 6) by incorporating Gaussian process surrogate models it can handle expensive simulators.    

In this paper we will develop POPE and apply it to two real-world cases: one fast stochastic simulator in the domain of stem cell biology and one slow deterministic simulator developed for cancer research.

\section*{Approximate Bayesian Computation}
The primary goal of Bayesian inference is to draw samples from or learn an approximate model of the following (usually intractable) posterior distribution:
\begin{equation}
  \pi(\thetav | \ystar_1, \ldots, \ystar_N ) \propto \pi(\thetav) \pi( \ystar_1, \ldots, \ystar_N  | \thetav )
\end{equation}
where $\pi(\thetav)$ is a prior distribution over parameters $\thetav \in {\rm I\!R}^{D}$ and $\pi( \ystar_1, \ldots, \ystar_N  | \thetav )$ is the likelihood of $N$ data observations, where $\ystar_n \in {\rm I\!R}^J$.  The vector of $J$ values can either be ``raw'' observations or, more typically, informative statistics of observations. In this paper we consider the case where $N=1$ (though all our methods apply equally to $N>1$) and will henceforth drop the subscripts.   The unconventional superscript on $\ystar$ is used to distinguish the observations from the simulator outputs $\y$. 

In ABC the likelihood function $\pi( \ystar | \thetav )$ is usually not available as a function but rather as a complex simulation, hence the alternative name for ABC, {\em likelihood-free inference}.  ABC sampling algorithms treat the simulator as an auxiliary variable generator and discrepancies between the simulator outputs and the observations as  proxies for the likelihood value. If we let $\y \simsim \pi( \y | \thetav )$ be a ``draw'' from the simulator, the likelihood can be written as:
\begin{equation}
  \pi( \ystar | \thetav ) = \int \lb \y = \ystar \rb \pi( \y | \thetav ) d\y \label{eq:abc_exact_likelihood}
\end{equation}
where $\lb \cdot \rb = 1$ if the arguments are true, and $0$ otherwise.  Equation~\ref{eq:abc_exact_likelihood} implies that we can compute the exact likelihood by integrating over all possible simulation output values.  In reality, since this integral requires simulations to match observations exactly, it is only achievable for discrete data. For continuous $\ystar$, $J$ slack variables $\epsvec$ are introduced around $\ystar$.  More specifically, an $\epsvec$-kernel function $\pi_{\epsvec}$ is used to measure the discrepancy between simulation results and observations. In practice the likelihood is approximated by a Monte Carlo estimate computed from $S$ draws of the simulator $\ymc \simsim \pi( \y | \thetav )$:
\begin{equation}
  \pi_{\epsvec}( \ystar | \thetav ) \approx  \int \pi_{\epsvec}(\ystar | \y ) \pi( \y | \thetav ) d\y 
                           \approx  \frac{1}{S} \sum_{s=1}^S \pi_{\epsvec}(\ystar | \ymc ) \label{eq:abc_mc_approx}
\end{equation}
This is clearly an \emph{unbiased} estimator of $\pi_{\epsvec}( \ystar | \thetav )$. Common $\pi_{\epsvec}$ functions are the $\epsvec$-tube $\pi_{\epsvec}(\ystar | \y ) \propto \prod_{j}\lb \| \ystarj-\yj \|_1 \leq \epsj \rb$ and the Gaussian kernel $\pi_{\epsvec}(\ystar | \y ) = \prod_{j}\mathcal{N} \lp \ystarj | \yj,  \epsj^2 \rp$.

Among the many possible ABC sampling algorithms, Markov chain Monte Carlo (MCMC) ABC is of particular relevance to this work \cite{marjoram2003markov,Wilkinson2013,sisson:2010}.  In the Metropolis-Hastings (MH) step the proposal distribution is composed of the product of the proposal for the parameters $\thetav$ and the proposal for the simulator outputs:
\begin{equation}
  q( \thetavp, \yp | \thetav ) =  q( \thetavp | \thetav ) \pi( \yp | \thetavp)
\end{equation}
i.e. parameters $\thetavp$ are first proposed, then outputs $\yp$ are generated from the simulator with input parameters $\thetavp$.

Using this form of the proposal distribution, and using the Monte Carlo approximation eq~\ref{eq:abc_mc_approx}, we arrive at the following Metropolis-Hastings accept-reject probability,
\begin{equation}
\alpha = \min \lp 1, \frac{\pi\lp\thetavp\rp \sum_{s=1}^S \pi_{\epsvec}(\ystar | \yps )  q( \thetav | \thetavp )}{\pi\lp\thetav\rp \sum_{s=1}^S \pi_{\epsvec}(\ystar | \ymc ) q( \thetavp | \thetav )} \rp \label{eq:abc_mh_acceptance_with_s}
\end{equation}
When only the numerator is re-estimated at every iteration (and the denominator is carried over from the previous iteration), then this algorithm corresponds to pseudo-marginal (PM) sampling \cite{delmoral2008,andrieu2009pseudo}. PM sampling is asymptotically correct (taking for granted the approximation introduced by the kernel $\pi_{\epsvec}$) but can display very poor mixing properties. By resampling the denominator as well, we improve mixing at the cost of introducing a further approximation. This sampler is known as the marginal sampler \cite{marjoram2003markov,sisson:2010}. Even the PM sampler requires $S$ simulations per MCMC move, which may be too expensive for complex simulators. Surrogate modeling---where the history of all simulations are stored in memory and used to build a surrogate of the simulator---may be the only option to make progress in that case.

\section*{Post Optimization Posterior Evaluation}
In regular ABC the simulator generates output statistics $\y$ that are  compared directly with observations $\ystar$.  For optimization problems, however, the scientist may interpret $\y_1$ as a cost and $\ystar_1$ as an estimate of the minimum cost.  Other simulation statistics $\{\y_j\}$, $j=2..J$ may  be constrained, e.g.  
 $\{\y_j \leq \ystar_j \}$. 
%
 For instance, the cost could be some measure of misfit between simulator outcomes and desirable outcomes while constraints could represent domains within which certain simulation results should lie (constraints can of course also be incorporated into the cost function, but as we will see, it is sometimes beneficial to treat them separately). Our first guess to elucidate some posterior distribution over parameters could be to define a Gibbs distribution $p(\y_1)\propto\exp(-\beta\y_1)$ which we would treat as a likelihood similar to $\pi_{\epsvec}$ and apply ABC, rejecting everything that does not satisfy the constraints. Unfortunately, we do not consider this a satisfactory solution because the posterior does not have a clear interpretation. For instance, simply scaling the arbitrary constant $\beta$ would change the posterior. 

A better solution is to define a new type of (one-sided) Heavyside kernel in ABC: $\lb \y_1 \leq \ystar_1 \rb$ which is $1$ when the argument is satisfied and $0$ otherwise. Note that this kernel is applied to both the objective $\y_1$ and the constraints $\{\y_j\}$ alike. The quantity $\ystar_1$ is given by the lowest value of the objective found by some optimization procedure (e.g. grid-search, Bayesian optimization \cite{snoek:2012}, etc). The posterior samples produced by an ABC algorithm that uses this one-sided kernel have a very clean interpretation, namely they represent \emph{the probability that a simulation run at that parameter value will generate an equally good or better (lower) value for the objective while satisfying all the constraints}. This distribution can be used to suggest new regions to explore (e.g. other modes, or regions that are farther away from constraint surfaces), and to visualize dependencies between parameters and their sensitivities. 

The posterior described above thus corresponds to
\begin{equation}
  \pi( \thetav | \ystar )  \propto  \pi(\thetav) \int \lb \y \leq \ystar \rb \pi( \y | \thetav ) d\y 
                           \propto  \pi(\thetav) \int_{-\infty}^{\ystar} \pi( \y | \thetav ) d\y 
                           \propto  \pi(\thetav) \F_{\y | \thetav}(\ystar) 
 \end{equation}
where $\F_{\y | \thetav}$ is the cumulative distribution function (CDF) of the conditional probability density function $ \pi( \y | \thetav )$ (or the probability of satisfying the constraint or improving the objective).  Since in ABC  we cannot compute the likelihood analytically, it is approximated by a Monte Carlo estimate:
\begin{equation}
   \F_{\y | \thetav}(\ystar) \approx  \frac{1}{S} \sum_{s=1}^S \lb \y^{(s)} \leq \ystar \rb ~~~~~~~~~~~~~~~~~~~~~~ \y^{(s)} \simsim \pi( \y | \thetav ) 
\end{equation}
Using the one-sided kernel $\lb \y \leq \ystar \rb$ will cause the ABC sampler to get stuck when initialized in a region where $\y > \ystar$ because every proposed sample will get rejected. Even when initialized in a region where $\y \leq \ystar$, this kernel will make it very difficult to move between different ``islands" (modes) in parameter space where these conditions hold. This problem is aggravated in high dimensions where $\lb \y \leq \ystar \rb = \prod_j \lb y_j \leq y_j^\star \rb $ and every condition needs to be satisfied for the likelihood to be non-zero.   A one-sided $\eps$-tube $\lb \y \leq \ystar + \eps \rb$ adds some relief but suffers the same problem for most useful values of $\eps$.   

The solution to this problem is to soften the kernel analogously to the softening of the condition $\lb \y = \ystar \rb$ into $\pi_{\epsvec}(\ystar|\y)$ in generalized ABC \cite{Wilkinson2013}.  If we define $d_j = \yj - \ystarj$, then these soft kernels treat all simulation outputs less than $\ystarj$ with likelihood proportional to $1$ and provide quadratic or linear penalties otherwise.   For example, a one-sided Gaussian kernel is defined as
\begin{align}
  K_{\eps_j}\lp \y_j;\; \ystar_j \rp = [ d_j \geq 0 ] +  [ d_j < 0 ] \exp\lp -\frac{1}{2} \lp\frac{d_j}{\eps_j}\rp^2\rp 
\end{align}
and a one-sided exponential kernel (i.e. linear penalty) is defined as
\begin{align}
  K_{\eps_j}\lp \y_j;\; \ystar_j \rp =  [ d_j \geq 0 ] +  [ d_j < 0 ] \exp\lp \frac{d_j}{\eps_j} \rp
\end{align}
 By modifying $\epsvec$ we can control the severity of the penalty, allowing us to use annealing schedules that adapt $\epsvec$ during the MCMC run in order to focus the sampling at modes when $\epsvec$ is small.  

Up to this point we have only discussed {\em one-sided} likelihoods, but there is nothing preventing the likelihoods to incorporate both upper and lower constraints:
\begin{equation}
   \pi\lp \ystar | \thetav \rp  =  \int_{\ystar_a}^{\ystar_b} \pi( \y | \thetav ) d\y 
                           =  \F_{\y | \thetav}(\ystar_b)-\F_{\y | \thetav}(\ystar_a)
\end{equation}
The one-sided kernels are easily modified for this, setting the likelihood to $1$ in between the regions, with quadratic or linear penalties outside of the regions.

\section*{MODELING THE SIMULATOR RESPONSE}
We may want to consider modeling the simulator response $\pi(\y|\thetav)$ if the outcome of the simulator is stochastic or the simulator is expensive to run. In the first case, we can reduce the variance of the Markov chain by learning a \emph{local response model} for every state  $\thetav$. For the second case, a {\em global response model} (a.k.a. a surrogate model) over the entire $\thetav$-space is more appropriate because it stores and makes use of the entire simulation history to predict responses at new $\thetav$ locations. 

\subsection*{Local Response Models}
When the simulator is fast and stochastic, it can be beneficial to the inference procedure to build a local, conditional model of the distribution $\pi(\y|\thetav)$ using $S$ simulator responses in $\yone, \ldots, \yS \overset{\simulator}{\sim} \pi( \y | \thetav)$.  
The simplest local response model is the \emph{conditional Gaussian}, an approach called {\em synthetic likelihood} in ABC \cite{wood2010statistical}.   It computes estimators of the first and second moments of the responses and uses the Gaussian distribution to analytically compute the likelihood (thus providing an alternative to kernel ABC). For our algorithms, this allows the direct computation of the CDF:
\begin{eqnarray}
 && \muhattheta  =  \frac{1}{S} \sum_{s=1}^{S} \y_s ~~~~~~~~~~~
 \hat{\Sigma}_{\thetav}  =  \frac{1}{S-1} \sum_{s=1}^{S} \lp \ys - \muhattheta \rp \lp \ys - \muhattheta \rp^T\\
 &&\F_{\y | \thetav}(\ystar; \muhattheta,  \hat{\Sigma}_{\thetav})  =  \int_{-\infty}^{\ystar} \mathcal{N}\lp \y | \muhattheta,  \hat{\Sigma}_{\thetav} \rp d \y
\end{eqnarray}
where $\muhattheta$ and $\hat{\Sigma}_{\thetav}$ are computed from the $S$ simulations.  In our experiments we  often use a factorized model: $\mathcal{N}(\y | \muhattheta,\hat{\Sigma}_{\thetav}) \approx \prod_{j=1}^J \mathcal{N}( y_j | \muhatthetaj,  \sigmahatthetaj)$, resulting in a factorized product over CDFs as well.   
Modeling the response by only the first two moments may be inadequate due to multi-modality, asymmetric noise, etc.   For such cases a \emph{conditional KDE} (kernel density estimate) response model can by used.  In \cite{TurnerGenLik2014} this approach is shown to be superior to conditional Gaussians for certain computational psychology models.  Note that for Gaussian kernels the conditional KDE is very similar to kernel ABC, but has additional flexibility of adaptively choosing bandwidths (rather than the fixed $\epsvec$ in kernel ABC).

\subsection*{Global Response Models}
For very expensive simulators it is impractical to run simulations at each parameter location in the MCMC run.  In these cases it is worth the extra storage and the computational overhead of learning a model of the simulator response surface.  For global response models the Metropolis-Hastings diverges from ABC-MCMC in that simulations are only performed if the surrogate is very uncertain.  When the surrogate is confident, no simulations are performed.

The natural global extension of the Gaussian conditional model is the Gaussian process (GP). The GP has been used extensively for surrogate modeling \cite{rasmussen:2003,kennedyohagan,lizotte2008,osborne2009}, including more recent applications in accelerating ABC \cite{Wilkinson2014,Meeds2014GpsUai}.  In \cite{Wilkinson2014} GPs directly model the log-likelihood in successive waves of inference, each one eliminating regions of low posterior probability.  This approach is capable of handling high-dimensional simulator outputs.   In \cite{Meeds2014GpsUai} each dimension of the simulator response is modeled by a GP and explicitly uses the surrogate uncertainty to determine simulation locations (design points).  The advantage of this approach is that CDFs can be computed directly from the GPs predictive distributions.  A global extension of the conditional KDE is more complicated, but  estimators such as the Nadayara-Watson could provide the necessary modeling machinery.  We leave these extensions to future work.

\begin{algorithm}[t] 
	\caption{POPE}
	\label{algo:kernelpope}
	\begin{algorithmic}[1]
  \Function{MCMC}{ $\thetav_0$, T, S, marginal, $\ystar$}
    \State $\thetav \gets \thetav_0$
    \State $\yone, \ldots, \yS \overset{\simulator}{\sim} \pi( \y | \thetav)$
    \For{$t = 1 : T$}
      \State $\thetapv \sim q(\thetapv | \thetav )$
      \State $\ypone, \ldots, \ypS \overset{\simulator}{\sim} \pi( \y | \thetavp)$
      \If{marginal}
        \State $\yone, \ldots, \yS \overset{\simulator}{\sim} \pi( \y | \thetav)$ \Comment{Marginal samplers do not keep simulations.}
      \EndIf
      \State $\alpha \gets \lp 1, \frac{\pi(\thetavp)q(\thetap | \thetavp ) \pi(\thetavp | \ystar, \epsvec)}{ \pi(\thetav) q(\thetapv | \thetap )\pi(\thetav | \ystar, \epsvec)} \rp$
      \If{$\mathcal{U}(0,1) < \alpha$}
        \State $\thetav \gets \thetavp$
        \State $\yone, \ldots, \yS \gets \ypone, \ldots, \ypS$
      \EndIf
      \State Collect $\thetav$\Comment{For posterior analysis.}
    \EndFor
     \State {\bf return} Collection $\thetav$
  \EndFunction
  \end{algorithmic}
\end{algorithm}

\section*{MCMC for POPE}\label{sec:adapting}

Algorithm~\ref{algo:kernelpope} provides the pseudo-code for running  kernel ABC POPE (easily modified to accommodate response models by plugging in the appropriate likelihood function).  This is simply ABC-MCMC with one-sided kernel likelihoods. There are two possible modes for running POPE: marginal and pseudo-marginal.  When running marginal MCMC, the state of the Markov chain only includes $\thetav$, and (as discussed earlier) has the property of improved mixing with the cost of doubling the number of simulations per Metropolis-Hastings step and a less accurate posterior.  On the other hand, pseudo-marginal can mix poorly, but uses fewer simulations and is more accurate.  Choosing between the two modes is problem specific.

\subsection*{Adaptive POPE}
In ABC, the choice of $\epsvec$ is crucial to both the MCMC mixing and the precision of the posterior distribution.  There is an obvious trade-off between the two as large $\epsvec$ provides better mixing but poorer approximations to the target distribution.  It is common in ABC to adapt $\epsvec$ using quantiles of the discrepancies (e.g. in Sequential Monte Carlo ABC \cite{beaumont2009adaptive}) or using a more complicated approach, for example based on the threshold acceptance curve \cite{Silk2013}.  We propose an online version of the quantile method (see function {\em UpdateEpsilons} in Algorithm~\ref{algo:adaptivepope} (Appendix~\ref{sec:appendixpope})), setting $\epsvec$ to a quantile of the exponential moving average (EMA) of the discrepancies or some minimum values $\epsmin$, which ever is greater.  Minimum values $\epsmin$ are set not only for computational reasons, but also to reflect the scientist's intuition regarding the relative importance of the constraints.  Because $\epsvec$ can fluctuate during the MCMC run, it can explore regions where some constraints are easily satisfied, but others are not, and vice-versa.  A quantile parameter $\epsquantile$ puts pressure on the chain to keep $\epsvec$ small.

For some problems we may not know certain  \emph{objective values} in $\ystar$ before running POPE.  For these cases simple adaptive MCMC procedures can estimate $\ystar$ during the MCMC run.  For deterministic simulators, $\ystar$ can be updated after each simulation.  For stochastic simulators we propose a local averaging procedure based on the EMA of $\y$, similar to the adaptation of $\epsvec$.  The intuition behind this is that the best objective value $\ystar$ at $\thetastar$ is the  expected value of the simulator response at $\thetastar$.  An EMA of the simulation response approximates this expectation and we have found in our experiments with stochastic simulators that it performs well and conveniently fits into the POPE MCMC procedure (i.e. there is no need to set up an entirely different optimization procedure with complicated constraints on the input and outputs since these are already part of POPE).  This is function {\em UpdateObjectives} in Algorithm~\ref{algo:adaptivepope} (Appendix~\ref{sec:appendixpope}).

These are adaptive MCMC algorithms that do not necessarily target the correct posterior distribution.  The simplest way to correct this is to simply use a few MCMC runs to set $\epsilon$ or $\ystar$ (if needed) or stop the adaptation altogether after a burnin period, from that point using non-adaptive ABC-MCMC.  An alternative to adapting $\epsilon$ is to include $\epsilon$ as part of the state of the Markov chain \cite{Bortot:2007}. 

\subsection*{Posterior Analysis of MCMC Results}
Along with the posterior parameter distribution $p( \thetav\,|\,\ystar)$, which is usually the main distribution of interest in a Bayesian analysis, we will also examine the {\em posterior predictive distribution}, denoted as $p(\y | \ystar)$, though perhaps unintuitive, is the distribution of statistics (the predictions) generated by the simulation at the parameters from $p( \thetav\,|\,\ystar)$.  Posterior predictive distributions are used in statistics for {\em model checking} and {\em model improvement} \cite{gelman}, for example, and use the generative model with parameters from the posterior to generate data, then statistics---defined by the statistician and considered important for the problem at hand---from the pseudo, or replicated data, are compared with the statistics from the observations (the real data).  One can then examine the bias and variance of the posterior predictive distributions with respect to the observations $\ystar$, or perform Bayesian t-tests (how probable are the observations $\ystar$ under $p(\y | \ystar)$) (see \cite{gelman}, Chapter XXX).

For ABC, the posterior analysis comes naturally, and usually, for free.  Using ABC-MCMC algorithms, statistics (judged important a priori by the scientist) are generated at each Metropolis-Hastings step.  Simply storing the pairs $\{\y, \thetav\}$ from the MH step is sufficient to produce both $p(\y\,|\,\ystar)$ and $p( \thetav\,|\,\ystar)$.  In addition to the posterior predictive, visualizing the input-output posteriors, i.e. a joint $p( \y_j, \thetav_d\,|\,\ystar)$ from the combined posterior predictive and posterior distribution, can lead to additional insight.

\section*{CASE 1: STEM-CELL NICHE GEOMETRY IN C. ELEGANS}
Minimizing the time it takes to develop an organ or to return to a desired steady state after perturbation is an important performance objective for biological systems \cite{Lander:2009fr,Itzkovitz:2012fj}. Control of the cycling speed of stem cells and of the timing of their differentiation is critical to optimize the dynamics of development and regeneration. This control is often exerted in part by stem cell niches. While stem cell niches are known to employ a number of molecular signals to communicate with stem cells \cite{Li:2005gz}, the impact of their geometry on stem cell behavior has received less attention. To begin to address this question, we ask here how niches should be shaped to minimize the amount of time to produce a given number of differentiated cells. 

We consider a model organ inspired from the C. \emph{elegans} germ line, which is similar to a number of other systems \cite{Cinquin:2009ep}. Cells reside within a tube-like structure; one end defined by the niche is closed, while the other is open and allows cells to exit. The set of possible positions that can be assumed by stem cells is constrained by the geometry of the niche; a dividing cell that is surrounded by neighbors pushes away one of its neighbors, which in turn might need to push away one of its own neighbors; cells pushed outside of the niche by one of these chain displacement reactions are forced to leave the cell cycle and differentiate. A simulator we developed tracks cell division and  movement, and outputs the time it takes to produce N cells for a given geometry. This geometry is such that rows are defined along the main axis of the organ; each cell row has its own size, comprised between 1 and 400 cells. There are several constraints that are put on the niche geometry to help the model remain realistic: the niche should hold fewer than 400 cells total, row size should monotonically increase along the niche axis, and the geometry should be ``well-behaved" (i.e., there should not be large jumps in row size along the axis).

\subsection*{Experimental set-up}
We performed several sets of experiments aimed at discovering the effects that realistic niche geometry constraints have on the time to 300 cells.  We therefore define a single statistic $\y_1$ to be the time to $N=300$ cells for a niche of $D$ rows; a niche geometry vector $\thetav$ defines the simulator input parameters.  In this study we set the number of rows in the niche to $D=8$.  To enforce the monotonicity constraints, we define $\thetav_1 = 1+ \geom_1$ and $\thetav_d = \thetav_{d-1}+\geom_d$, $\forall d > 1$, i.e. we define niche geometries in terms of niche increment parameters $\geom_d \geq 0$.  With this set-up, we can change the prior constraints and observe the effects on the posterior predictive distribution $p(\y_1 | \ystar_1)$.  

There are three sets of constraints on $\thetav$ (and/or $\geom$), each with their own kernel epsilon parameter; the constraint $\geom_d \geq 0$ is strictly enforced.  For all experiments, the first cell row was given a flexible range $\thetav_1 \in \{1,400\}$, thus the first constraint is $K_{\eps_{g_1}}\lp \geom_1;\; \tau_{g_1} \rp$, where $\eps_{g_1}=0.1$ and $\tau_{g_1} = 399$.  
The second set of constraints is on the niche geometry increments $K_{\eps_{g_d}}\lp \geom_d;\; \tau_{g_d} \rp $, where $\eps_{g_d} = 0.1$ and $\tau_{g_d}$ is set to $10$ (to capture well-behaved niche increments) or $399$ (essentially removing the constraint on niche increments); see experiment details below.  The final constraint on $\thetav$ is on the total niche geometry size $K_{\eps_{\theta}}\lp \sum_{d=1}^D \thetav_d;\; \tau_{\theta} \rp$, where $\eps_{\theta} = 1$ and $\tau_{\theta}$ is set to $400$ or $1500$.  For all experiments, a one-sided Gaussian kernel was used.  The prior over $\geom$ is therefore:
\begin{align*}
  \pi\lp\geom \rp &\propto  K_{\eps_{\theta}}\lp \sum_{d=1}^D \thetav_d;\; \tau_{\theta} \rp K_{\eps_{g_1}}\lp \geom_1;\; \tau_{g_1} \rp  \prod_{d=2}^D K_{\eps_{g_d}}\lp \geom_d;\; \tau_{g_d} \rp 
\end{align*}

The likelihood is a one-sided kernel $\pi\lp \ystar\,|\,\y_1 \rp \propto K_{\eps_y}\lp \y_1;\; \ystar_1 \rp$, where $\eps_y = 0.01$ (except for experiment D, below) and $\ystar_1 = 27.05$.  For this problem we did not know $\ystar_1$ a priori, so we ran 5 runs of marginal kernel ABC with $S=1$ and adapted $\ystar$ (see algorithm 2 in appendix). We choose $\ystar_1 = 27.05$, the median value from 5 runs (which produced values $26.99$, $27.03$, $27.05$, $27.07$, $27.28$).  Table~\ref{tab:niche_parameters} summarizes the parameters and results from these experiments.  For all experiments, $5$ marginal ABC-MCMC runs of length $10000$ were run and the first $2000$ samples were discarded as burnin.  

\begin{table}[h]
  \centering
  \begin{tabular}{|c||c|c|ccc|ccc|c|}
    \hline
    Experiment & M  & $\ystar$   & $\tau_{g_1}$ & $\tau_{g_d}$   & $\tau_{\theta}$   & $\mean{\y_1}$ & $\median{\y_1}$ & $\mode{\y_1}$ & $P( \y_1 < 27.05)$\\ \hline \hline
 \multirow{2}{*}{A} &  1 & 27.05     & $399$ & $399$ & $400$ & 27.042 & 27.037 & 27.029 & 0.53 \\
                    &  1 & 27.05     & $399$ & $10$  & $400$ & 27.059 & 27.054 & 27.076 & 0.49 \\ \hline
\multirow{2}{*}{B}  &  1 & $\infty$    & $399$ & $399$ & $400$ & 27.078 & 27.081 & 27.076 & 0.43 \\ 
                    &  1 & $\infty$    & $399$ & $10$  & $400$  & 27.298 & 27.150 & 27.114  & 0.32 \\ \hline
\multirow{2}{*}{C}  &  1 & $\infty$    & $399$ & $399$ & $1500$ & 30.159 & 30.184 & 30.224 & 0.00  \\ 
                    &  1 & 27.05    & $399$ & $399$ & $1500$ & 27.322 & 27.227 & 27.150 & 0.24 \\ \hline
\multirow{2}{*}{D}  & 10 & 27.05    & $399$ & $399$ & $400$ & 27.053 & 27.049 & 27.043 & 0.51 \\
                    & 10 & 27.05    & $399$ & $10$  & $400$ & 27.056 & 27.053 & 27.050 & 0.47  \\  \hline
  \end{tabular}\caption{\small{Stem-cell niche geometry experimental set-up and posterior predictive results.  M is the number of replicates used to compute a statistics (see Experiment D).  See text for definitions of other columns.}}
  \label{tab:niche_parameters}
\end{table}

\subsection*{Experiment A: realistic constraints on $\geom_d$}

\begin{figure}[h]
  \begin{center}
    \includegraphics[width=0.8\columnwidth]{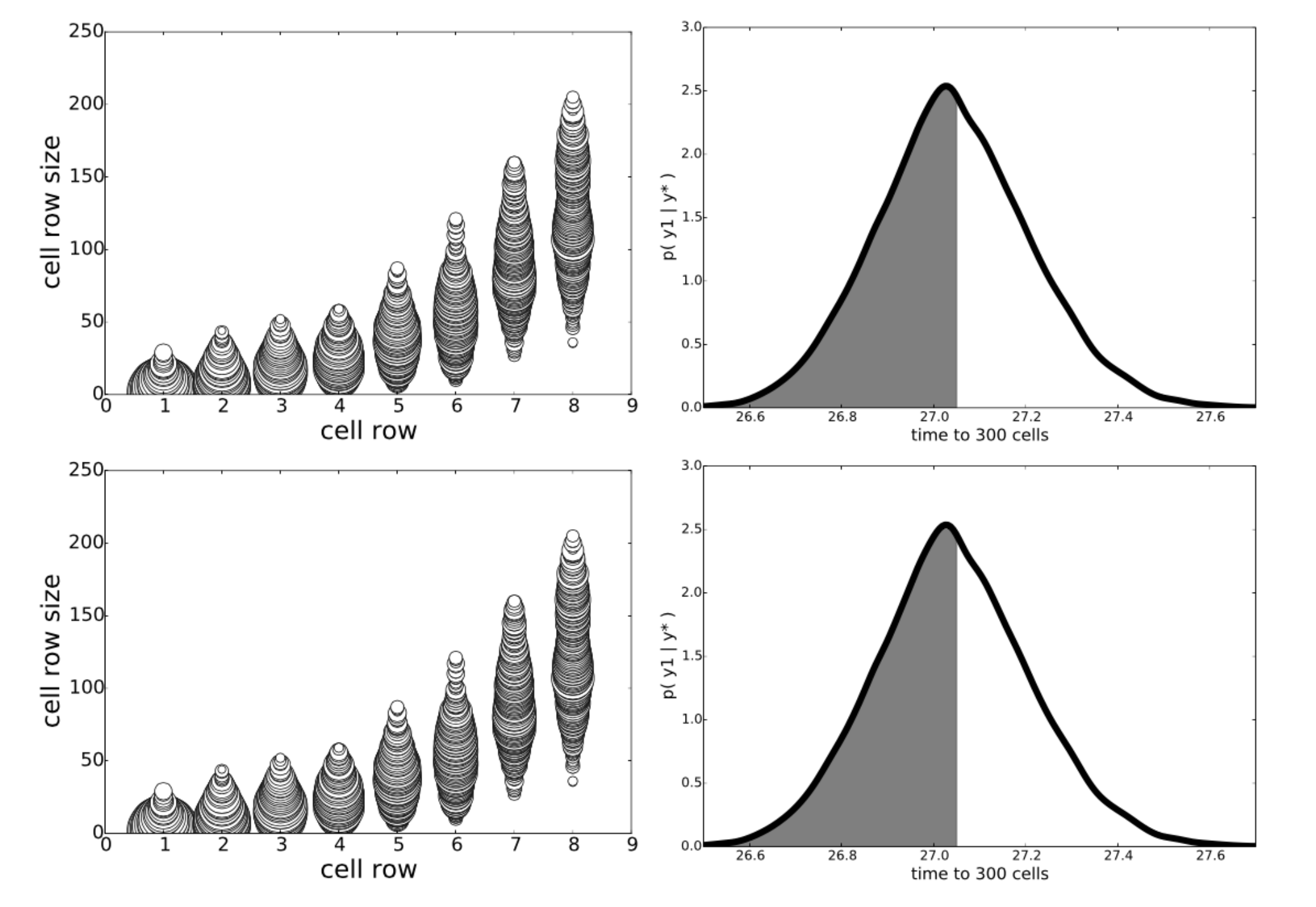}
\end{center}
\vspace{-0.2in}
\caption{\small{Stem-cell niche geometries, Experiment A}.  Comparison of niche geometry posteriors with $\tau_{g_d} = 400$ (top row) and $\tau_{g_d} = 10$ (bottom row).  The left column illustrates the posterior geometries $\thetav$ by plotting circles of radius proportional to their posterior fraction of that size for that row (rounded to integers).  The right column is the posterior predictive distribution $p(\y_1 | \ystar_1)$, with shading indicating the probability mass $P(\y_1 < 27.05\,|\,\ystar_1)$.  }
\label{fig:nicheA}
\end{figure}

The first set of experiments compared posterior inference using a $\tau_{g_d}=399$ and $\tau_{g_d}=10$.  
Figure~\ref{fig:nicheA} shows the posterior geometries with $\tau_{g_d}=399$ (top row) and with a realistic constraint $\tau_{g_d}=10$ (bottom).  Without the realistic constraint, the sizes start smaller (averaging around 5), increase slowly until row 6, then jump to a larger size (over 100) at row 8.   With the realistic constraint, the sizes start larger (averaging around 20), and increase steadily until row 8, with no jumps, to an average of about 50.  The posterior predictive distributions are very similar for both results, with the probability of $\y_1 < 27.05$ without the constraint being $0.53$ compared to $0.49$ with the constraint, indicating that the constraints do remove some regions of the parameter space with shorter time to 300 cells.  The medians and modes of $\y_1 | \ystar_1$ also support this (without: $27.037$/$27.029$, with: $27.054$/$27.076$).
\def \thisscale{0.35}

\subsection*{Experiment B: removing constraint on time to 300 cells}
\begin{figure}[h]
  \begin{center}
    \includegraphics[width=0.8\columnwidth]{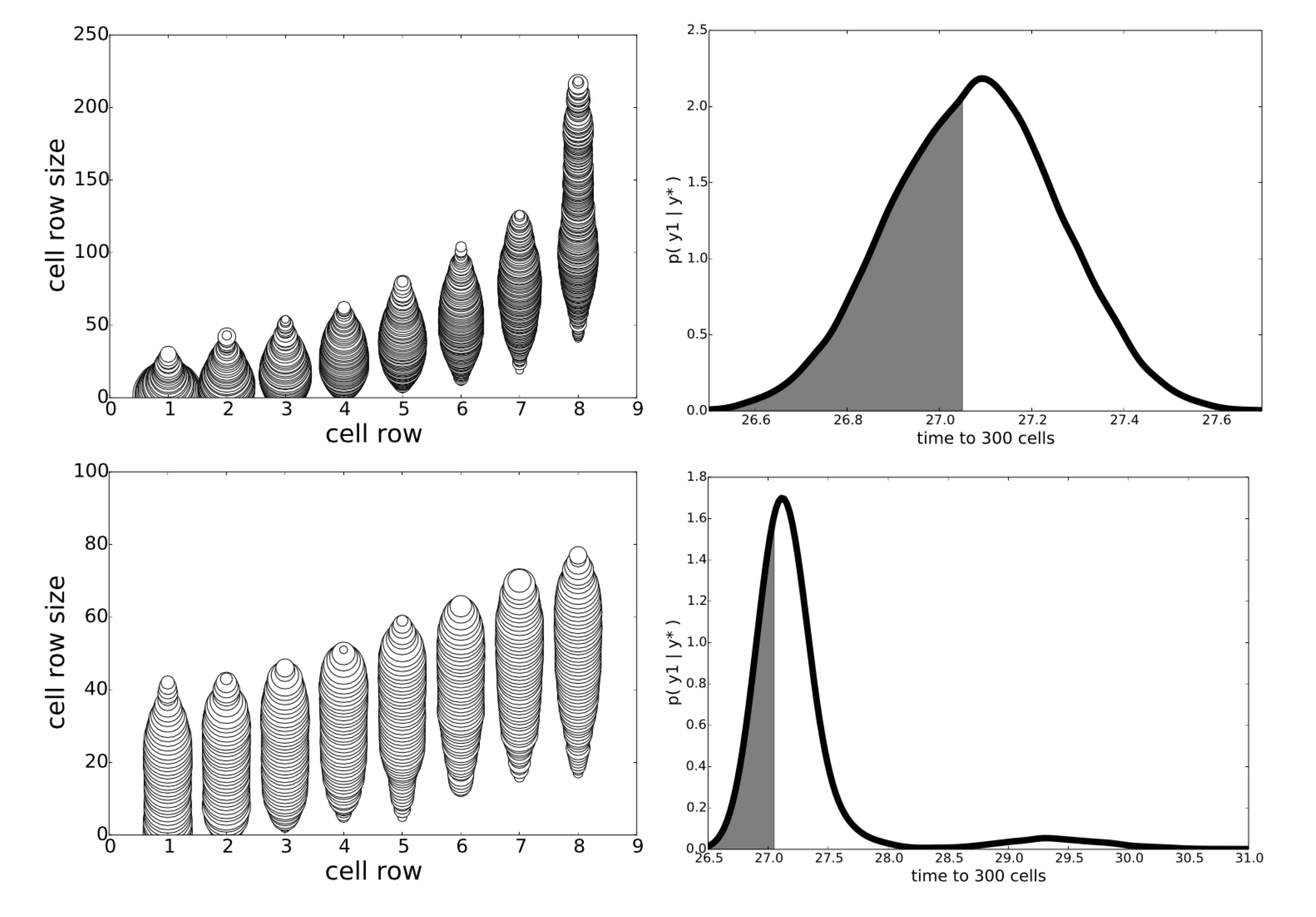}
\end{center}
\vspace{-0.2in}
\caption{\small{Stem-cell niche geometries, experiment B. Comparison of niche geometry posteriors with $\tau_{g_d} = 400$ (top row) and $\tau_{g_d} = 10$ (bottom row), but with the likelihood term removed.}}
\label{fig:nicheB}
\end{figure}

We next removed the effect of the likelihood term on the posterior by setting $\ystar_1 = \infty$ (which is equivalent to sampling from the prior, with soft boundaries, using MCMC).  Results for this experiment are shown in Figure~\ref{fig:nicheB}.  Surprisingly, the posteriors of $\thetav$ have the same form as in experiment A, though with some decreases in $P(\y_1 < 27.05\,|\,\ystar_1)$:  from $0.53$ to $0.43$ (for $\tau_{g_d}=399$) and from $0.49$ to $0.32$ (for $\tau_{g_d}=10$).  This result clearly shows that there is significant {\em prior mass} having $\y_1 < 27.05$.  
\def \thisscale{0.35}
 
\subsection*{Experiment C: increasing threshold on total niche cells}
\begin{figure}[h]
  \begin{center}
    \includegraphics[width=0.8\columnwidth]{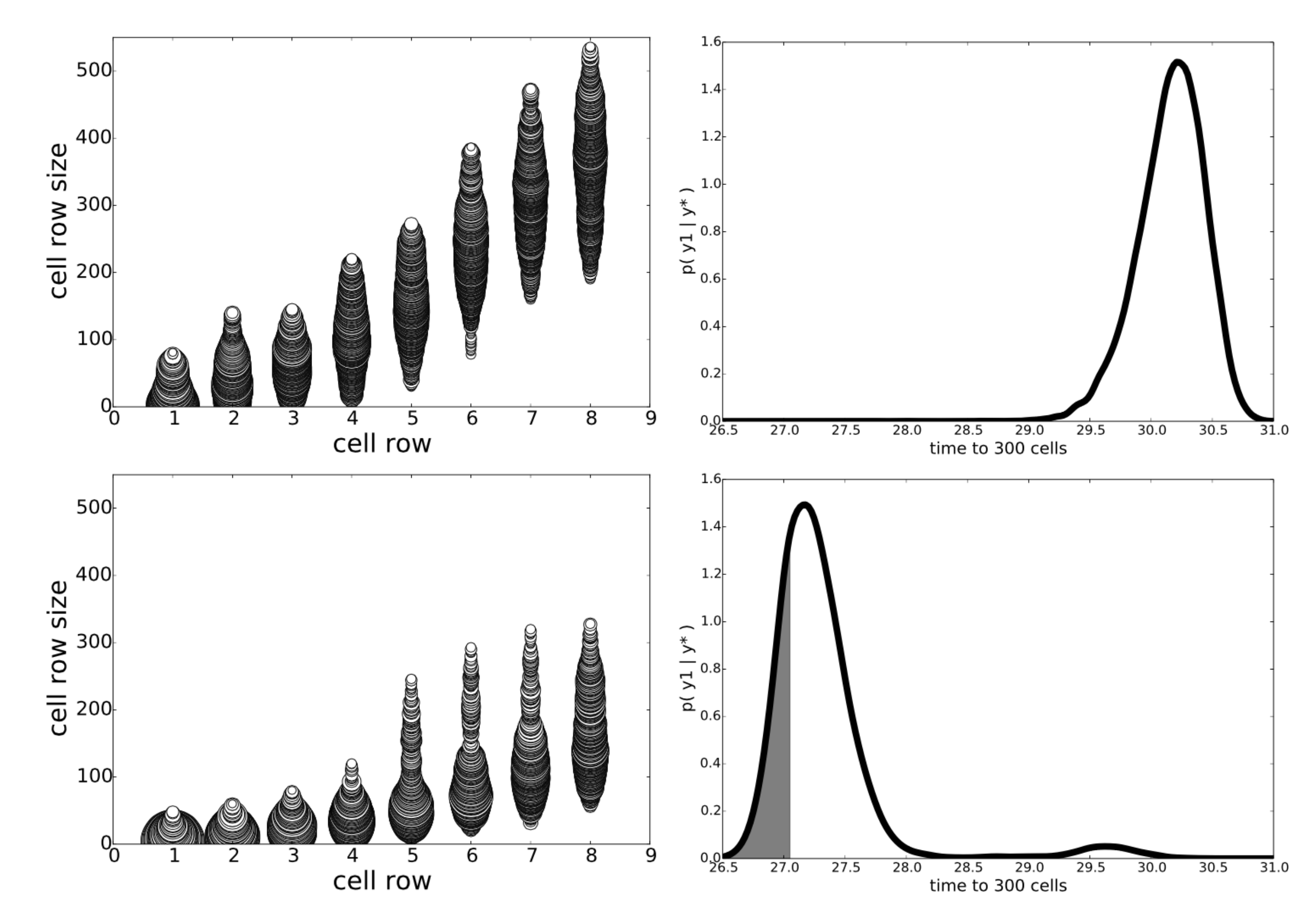}
\end{center}
\vspace{-0.2in}
\caption{\small{Stem-cell niche geometries, experiment C.  In these experiments, $\tau_{\theta} = 1500$.  Top: $\ystar_1 = \infty$.  Bottom: $\ystar_1=27.05$.  When $\ystar_1=27.05$, posteriors are similar to experiment A (Figure~\ref{fig:nicheA}, top).}}
\label{fig:nicheC}
\end{figure}

In this experiment we compare an increase in $\tau_{\theta}$ in an attempt to determine the most important factor for minimizing the time to 300 cells, the likelihood constraint $\ystar_1$ or the constraint on the total size.  Results are shown in Figure~\ref{fig:nicheC}.  For both results, $\tau_{\theta} = 1500$, but in the top row, the likelihood constraint is removed (and kept in the bottom row).  By increasing the total  niche geometry permitted ($\tau_{\theta} = 1500$) and removing the constraint on $\y_1$ (top row), the posterior predictive distribution degrades severely, with no samples satisfying $\y_1 < 27.05$.  However, when the constraint on $\y_1$ is reintroduced (bottom row), a $P(\y_1 < 27.05\,|\,\ystar_1)=0.24$ is significant, and the posteriors of $\thetav$ are very similar to experiment A with $\tau_{g_d}=399$.
\def \thisscale{0.35}

The results of experiments A-C demonstrate the relative importance of the input and output constraints on the posterior probability of $\y_1\,|\,\ystar_1$.  The most important constraints are $\sum \theta_d$  and $\y_1 < \ystar_1$.   Both have similar effects on the posterior predictive distribution.  The constraint $\tau_{g_d}$ has little effect on $P(\y_1 < 27.05\,|\,\ystar_1 )$, but does produce significantly different posterior geometries, mainly due to the prior constraints. 

\subsection*{Experiment D: replacing statistics with average of replicates}

\begin{figure}[h]
 \begin{center}
   \includegraphics[width=0.8\columnwidth]{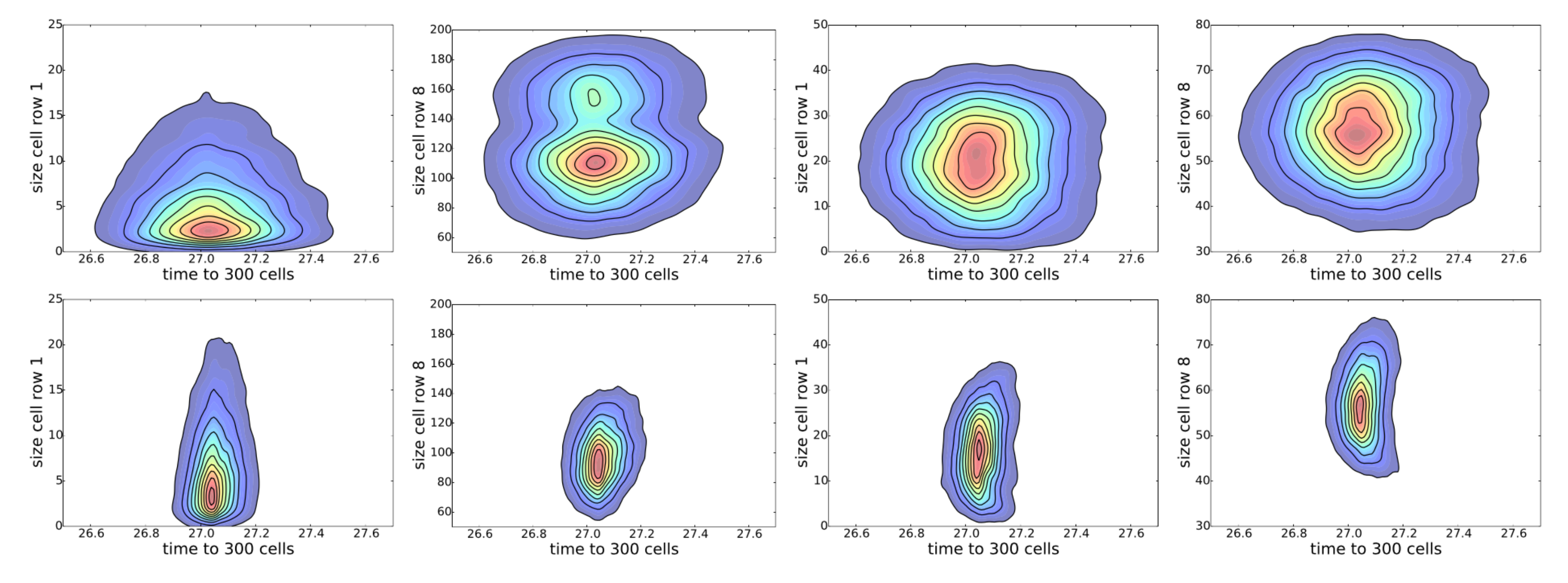}
 \vspace{-0.2in}
 \end{center}
 \caption{\small{Stem-cell niche geometries, experiment D.  Effect of $M$, the number of replicates used to compute the output statistic $\y_1$:  $M=1$ (top) versus $M=10$ (bottom).  The left 2 columns correspond to $\tau_{g_d}=399$ and the right 2 columns $\tau_{g_d}=10$.  Each plot is a joint posterior $p(\y_1, \theta_d\,|\,\ystar)$, for $d\in\{1,8\}$.}}
 \label{fig:nicheD}
 \end{figure}

One final experiment on niche geometries was performed, aimed at exploring the effect that reducing the simulator noise has on the posterior distributions.  To do this, we repeat each simulation $M$ times, using the same parameter setting; i.e. $\y = \frac{1}{M} \sum_{m=1}^M \y^{(m)}$, where $\y^{(m)} \simsim \pi( \y | \thetav)$. The variance of the statistic therefore decreases with $M$.  Although, as expected, the posterior predictive distribution contracts around $\y$, we found no significant changes to the posterior $p(\thetav | \ystar)$ when $M=1$ (see Figure~\ref{fig:nicheD}).  This experiment gives evidence that the scientist should instead change the value $\eps$ to control the posterior predictive distribution rather than $M$, which has an $M$-fold increase in computation.   \\
 %
 \def \thisscale{0.22}
 \vspace{-0.1in}
 \def \thisscale{0.35}
 \begin{figure}[h]
   \begin{center}
     \includegraphics[width=0.8\columnwidth]{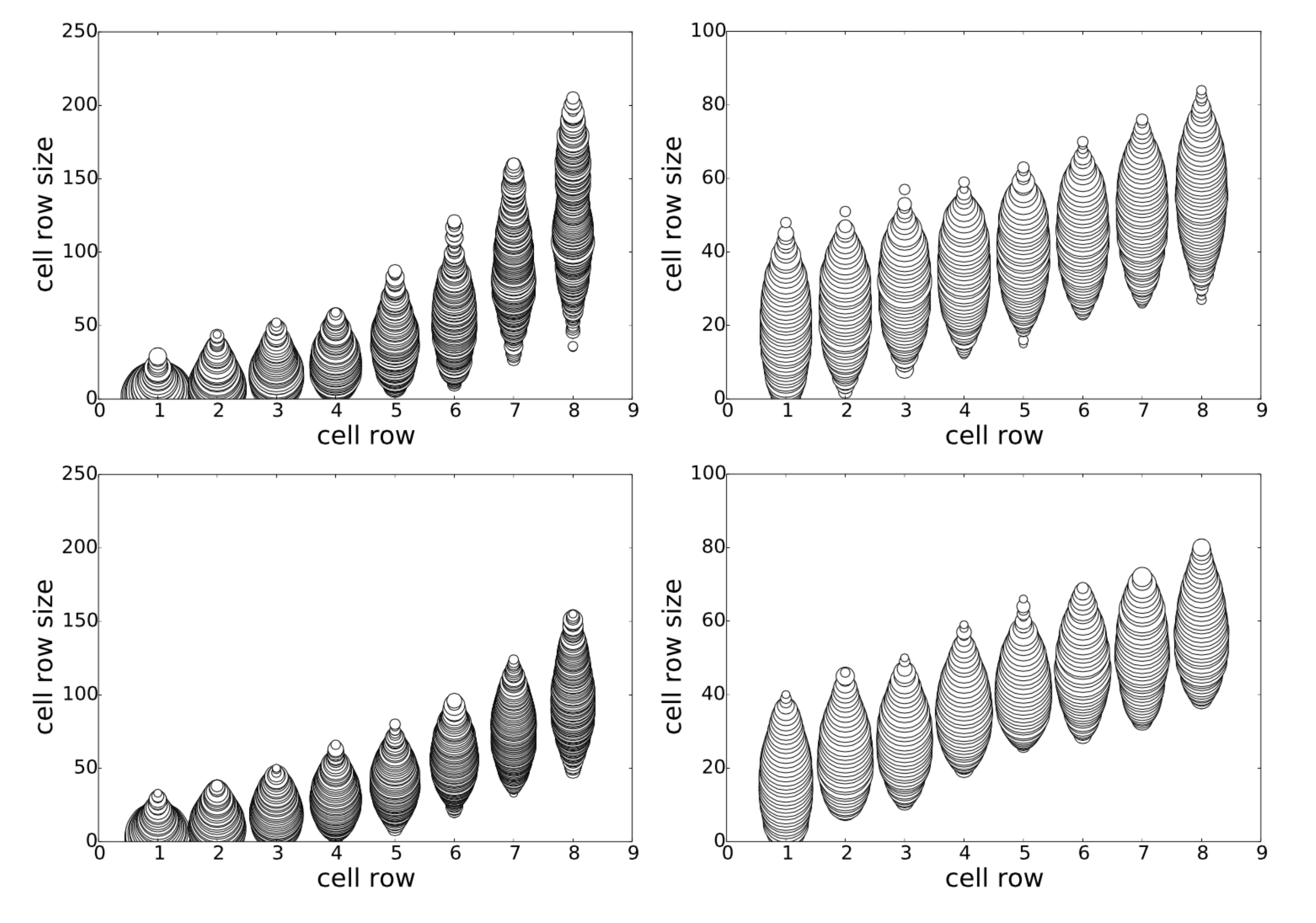}
     \vspace{-0.2in}
 \end{center}
 \caption{\small{Stem-cell niche geometries, Experiment D}.  Effect of $M$, the number of replicates used to compute the output statistic $\y_1$:  $M=1$ (top) versus $M=10$ (bottom).  The left column correspond to $\tau_{g_d}=399$ and the right column $\tau_{g_d}=10$.  }
 \label{fig:nicheDb}
 \end{figure}

 Experiments A-D illustrate the usefulness of POPE for exploring the roles constraints play on the optimization posterior.  We found that the constraints on the prior over valid regions of $\thetav$ had significant influence on the posterior, and played a similar role to the likelihood term.  Using realistic constraints on changes in row sizes had very little detrimental effect on the time to 300 cells, compared to having no realistic constraint.  More important was the  constraint on total geometry size.  We found very little difference in the posteriors when the statistics were averages of simulation replicates versus a single simulation.  This makes sense if the simulation noise is taken into account when setting $\epsilon$: when increasing the number of replicates in the average, $\epsilon$ should be decreased (from its setting at $M=1$) to take into account the population mean variance, but this seems unnecessary since the posteriors change little, but the number of simulations increases.

\clearpage
\section*{CASE 2: SPOTTED PATTERNS IN COLON CANCER TUMORS}

A remarkable pattern of spots is visible in the tissue of colon cancer tumors when stained for markers indicating glycolytic activity.  It is hypothesized that the spotted regions indicate localized areas of glycolytic cells, whereas surrounding areas are considered oxidative cells. Furthermore it is thought that Wnt signaling (an important cell signaling pathway in development and healing) plays a critical role in reducing glycolytic activity \cite{Pate2014}, thereby resulting in significant changes in spot formation. Experiments blocking Wnt by overexpression of a dominant negative form of lymphoid enhance factor (dnLEF-1) have shown that interfering with the Wnt pathway leads to fewer but larger spots and lighter background staining color than \emph{Mock} tissue (tumors that have not received dnLEF-1 intervention).  

Based on these findings, a simulator of a mathematical model of reaction-diffusion equations was built that produces spatial and temporal dynamics of a population fraction of oxidative cells and glycolytic cells, as well as the activity of Wnt and a Wnt inhibitor. The Wnt and Wnt inhibitor equations are based on the Gierer-Meinhardt activator-inhibitor model, where Wnt is the activator which produces a factor that inhibits Wnt activity.  

\begin{figure}[b]
\captionsetup[subfigure]{labelformat=empty, labelsep=none}
\begin{center}
  \includegraphics[width=\columnwidth]{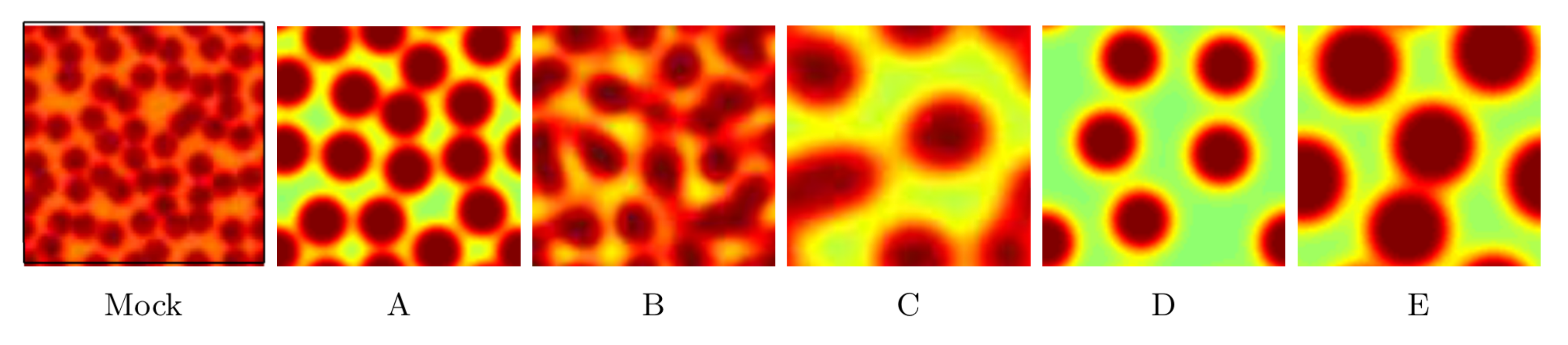}
\vspace{-0.2in}
\caption{\small{2D simulation patterns of glycolic cells at the final time step.  See text for details.}}
\label{fig:spots_at_fifty}
\end{center}
\end{figure}

The goal of these experiments is to provide feedback to the mathematical biologists regarding the characteristics of simulation parameters that produce {\em simulated patterns different from Mock patterns}.  For this reason, this problem does not have a predefined cost function, but instead uses the observed Mock values as constraints.  
The simulation produces 1D spatial and temporal patterns (see Figure~\ref{fig:spots_at_fifty} for 2D examples) from which $J=4$ statistics are computed: $\y_1$ the average spot width (based on wave patterns in 1D images); $\y_2$ the number of spots (waves, in 1D); $\y_3$ the average background level; and $\y_4$ the average Wnt level.  There are $D=9$ simulator parameters including rates of production and decay for Wnt and Wnt inhibitor, and their diffusion coefficients. These are described in Table~\ref{table:spottheta}.  The $\thetav$ settings in column {\em Mock} in Table~\ref{table:spottheta} generate patterns that were judged similar to the Mock spotting patterns in tissue photographs.  Their corresponding statistics $\ystar = \{0.604, 5, 0.807, 5.67\}$ are shown in Table~\ref{table:spotstats}, along with statistics from other $\thetav$ settings $A$ to $E$, described below.

The Mock values $\ystar$ define the constraints on simulator statistics $\y$.  More precisely, they constrain the posterior to regions where 
 $\lb \y_1 > \ystar_1\rb$, $\lb \y_2 < \ystar_2\rb$, $\lb \y_3 < \ystar_3 \rb$, and $\lb \y_4 < \ystar_4 \rb$,  
 which correspond to the goal of producing different patterns from Mock.  For example, the first constraint states that we want the spot widths from simulation to be greater than $\ystar_1 = 0.604$, the average width of spots for the Mock setting $\thetav$.  Similarly, we want fewer than $5$ spots, a background lighter than $0.807$, and a Wnt level less than $5.67$.  Further constraints are added to avoid degenerate simulation results; as an example, we  set its likelihood to zero when there are no spots detected.  

\def \thisscale{0.15}

 \begin{table}[h]
 \caption{Simulation parameters $\thetav$ for spotted patterns in colon cancer tumors.}
 \label{table:spottheta}
 \begin{center}
 \begin{tabular}{|l|l|l||l|l|l|l|l|}
 \hline
 Parameter $\thetav$ & Description & Mock & A & B & C & D & E\\
 \hline
 $\kappa_W > 0$     & Rate of nonlinear Wnt production        & $4$ &  $0.442$ & $0.951$ & $2.44$ & $0.399$ & $0.315$       \\
 $\kappa_{W_I} > 0$ & Rate of Wnt inhibitor production        & $1$ & $27.4$ & $0.484$ & $0.161$ & $0.486$ & $0.188$         \\
 $\mu_W \geq 0$        & Decay rate of Wnt                    & $2$ & $0.642$ & $0.179$ & $0.791$ & $0.545$ & $0.936$       \\
 $\mu_{W_I} \geq 0$    & Decay rate of Wnt inhibitor          & $4$ & $2.36$ & $1.30$ & $1.10$ & $0.569$ & $1.064$     \\
 $a \geq 0$            & Constant of inhibition               & $10^{-8}$ & $0.4006$ & $0.416$ & $0.0384$ & $0.00491$ & $0.0284$ \\
 $b \geq 0$            & Constant of inhibition by $W_I$      & $1$ & $0.0125$ & $7.94$ & $20.05$ & $0.616$ & $0.640$   \\
 $S_W \geq 0$          & Rate of constitutive  Wnt production & $1$ & $0.00167$ & $ 0.00351$ & $17.75$ & $0.00005$ & $0.00009$  \\
 $1 \geq D_W > 0$          & Diffusion coefficient of Wnt     & $0.01$ & $0.0180$ & $0.00322$ & $0.0955$ & $0.0336$ & $0.0810$  \\
 $1 \geq N > 0$            & Nutrient level                   & $1$ & $0.818$ & $0.897$ & $0.984$ & $0.959$ & $0.970$   \\
 \hline
 \end{tabular}
 \end{center}
 \end{table}
  
 \begin{table}[h]
 \caption{Simulation statistics $\y$ for spotted patterns in colon cancer tumors.}
 \label{table:spotstats}
 \begin{center}
 \begin{tabular}{|l|l|l||l|l|l|l||l|}
 \hline
 Statistic $\y$ & Feasible Region & Mock ($\ystar$) & A & B & C & D & E\\
 \hline
 Avg. Spot Width & $\y_1 > 0.604$             & $0.604$ & $1$    & $0.65$  & $0.65$ & $1$ & $1.75$ \\
 Number of Spots & $\y_2 \in \lb 2, 3, 4 \rb$ & $5$     & $3$    & $4$   & $2$    & $3$ & $2$    \\
 Avg. Background & $\y_3 < 0.807$             & $0.807$ & $0.77$ & $0.75$  & $0.70$ & $0.6$ & $0.70$\\
 Avg. Wnt        & $\y_4 < 5.67 $             & $5.67$  & $3.25$ & $1.50$  & $0.75$ & $1$ & $2$\\
 \hline
 \end{tabular}
 \end{center}
 \end{table}

This simulator is deterministic but expensive to evaluate, requiring roughly 30 seconds to complete for the 1D simulator used in our experiments, and 90 seconds for the 2D simulator, used for generating 2D images only.  We ran $6$ chains of length $4000$ pseudo-marginal kernel ABC-MCMC with S=1.  To initialize the chains, a short rejection sampling procedure was used to select $\thetav_0$ for each random seed.  This is necessary as many random configurations of $\thetav$ result in degenerate simulation results (i.e. zero likelihood).  Diffuse log-normal prior distributions were placed over $\thetav_1$ to $\thetav_7$ and weak Beta priors put on $D_{W}$ and $N$. At least $100$ initial samples were discarded from each chain; sometimes more if the chain had not yet reached a location where all the constraints were satisfied.  In total there were $22257$ samples in the posterior.

\begin{figure}[h]
\begin{center}
  \includegraphics[width=\columnwidth]{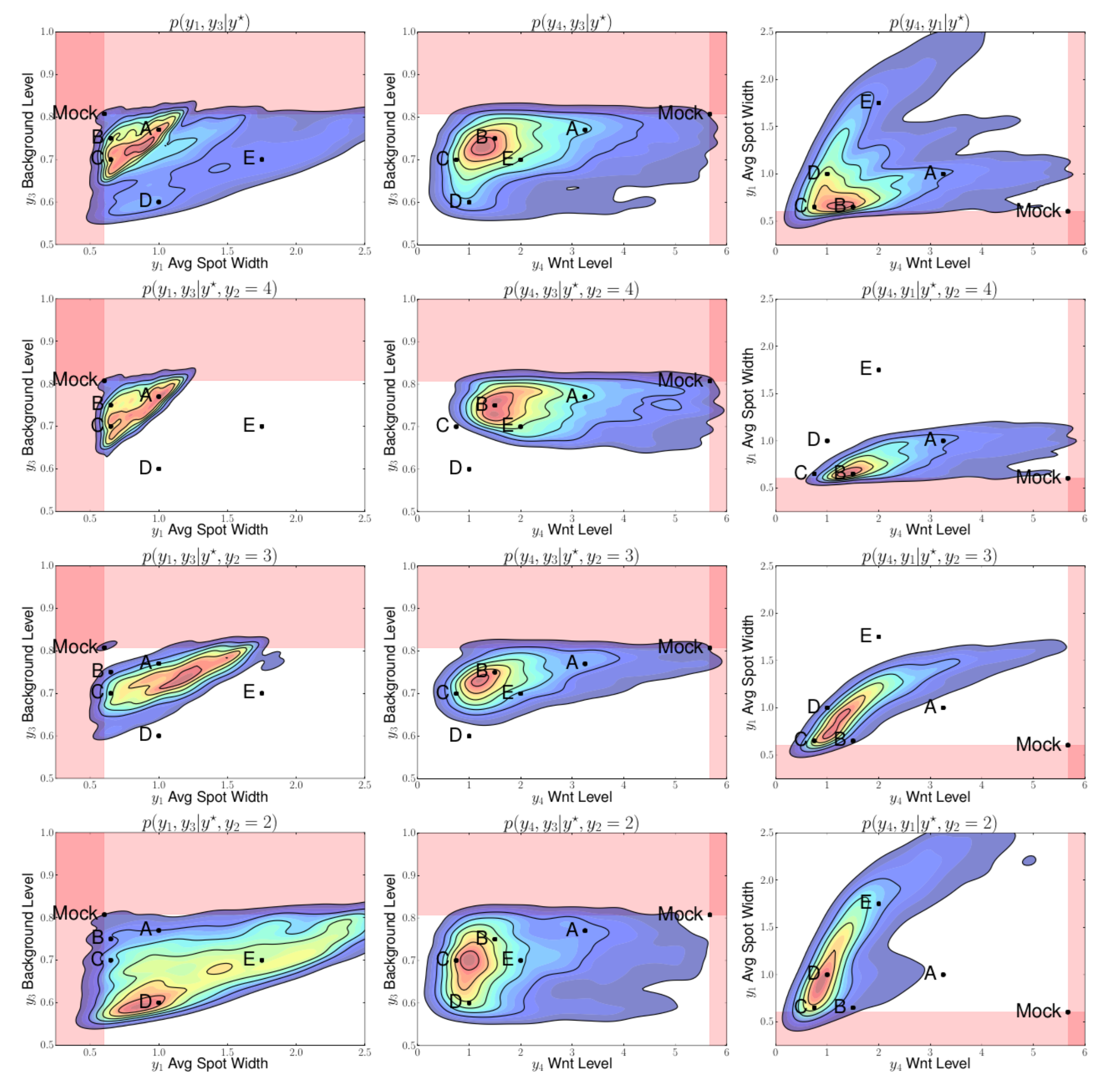}
\end{center}
\vspace{-0.2in}
\caption{\small{Posterior predictive distributions (PPDs) shown marginally for pairs of statistics.  {\bf Row 1:} The full PPD. {\bf Rows 2 to 4:} spot-conditional PPDs for spot numbers 4 to 2, respectively.  Columns differ on pairs of statistics.  Mock constraints indicate invalid regions in shaded pink.  Spot-conditional modes are indicated by letters $A$, $B$, and $C$ for spots 4 to 2, along with $D$, a statistic deemed {\em far away} from Mock, by visual inspection.}}
\label{fig:spots_ppd_pairwise}
\end{figure}

\begin{figure}[h]
\begin{center}
  \includegraphics[width=\columnwidth]{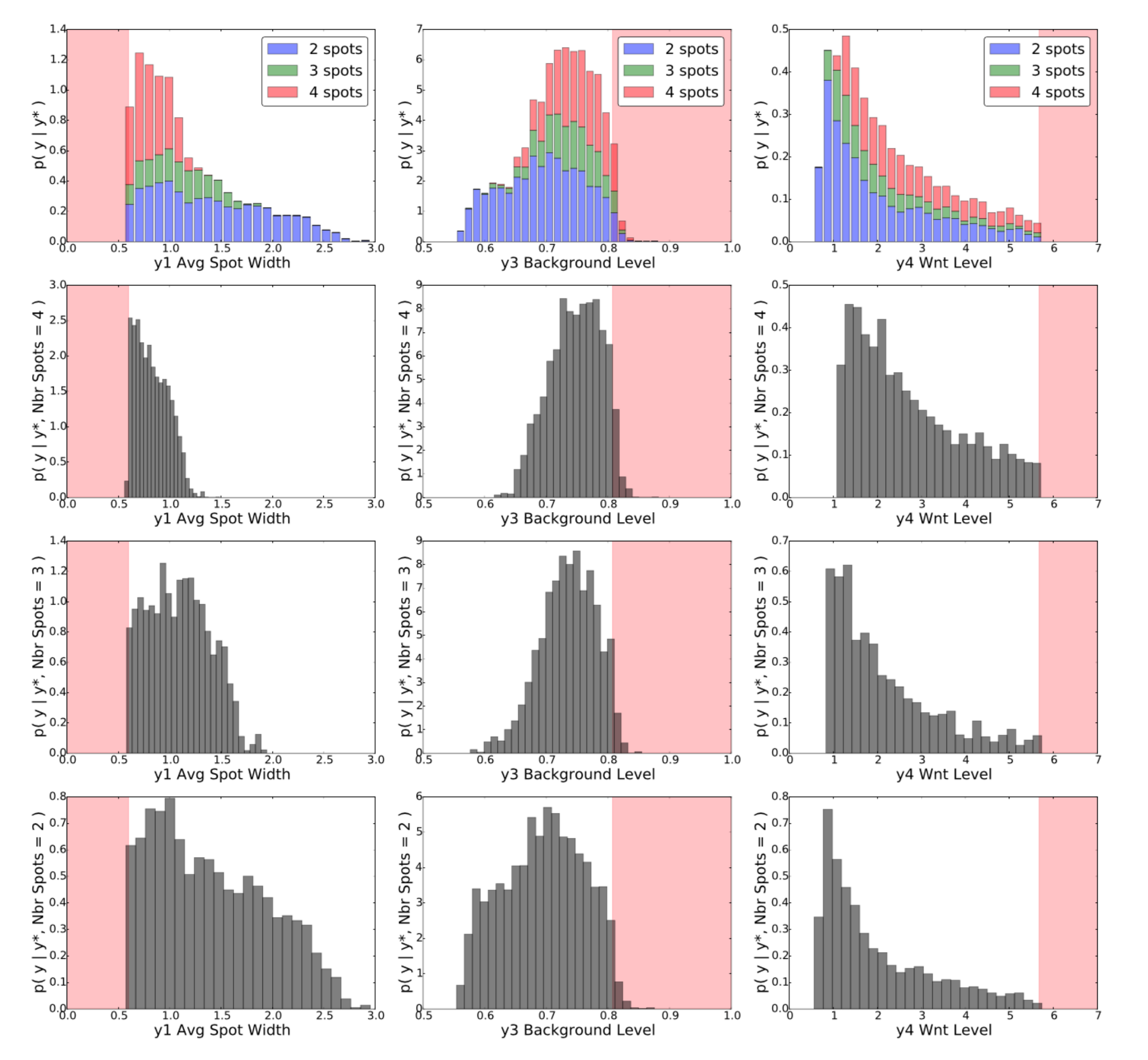}
\end{center}
\vspace{-0.2in}
\caption{\small{Marginal posterior predictive distributions (PPDs); one column for statistics $\y_1$, $\y_3$, and $\y_4$.  {\bf Row 1:} The full PPDs as histograms, but using colors {\em blue} ($\y_2 = 4$), {\em green} ($\y_2 = 3$), and {\em red} ($\y_2 = 2$) to differentiate spot numbers . {\bf Rows 2 to 4:} spot-conditional PPDs for spot numbers 4 to 2, respectively.}}
\label{fig:spots_ppd_marginal}
\end{figure}

\begin{figure}[t]
\begin{center}
  \includegraphics[width=\columnwidth]{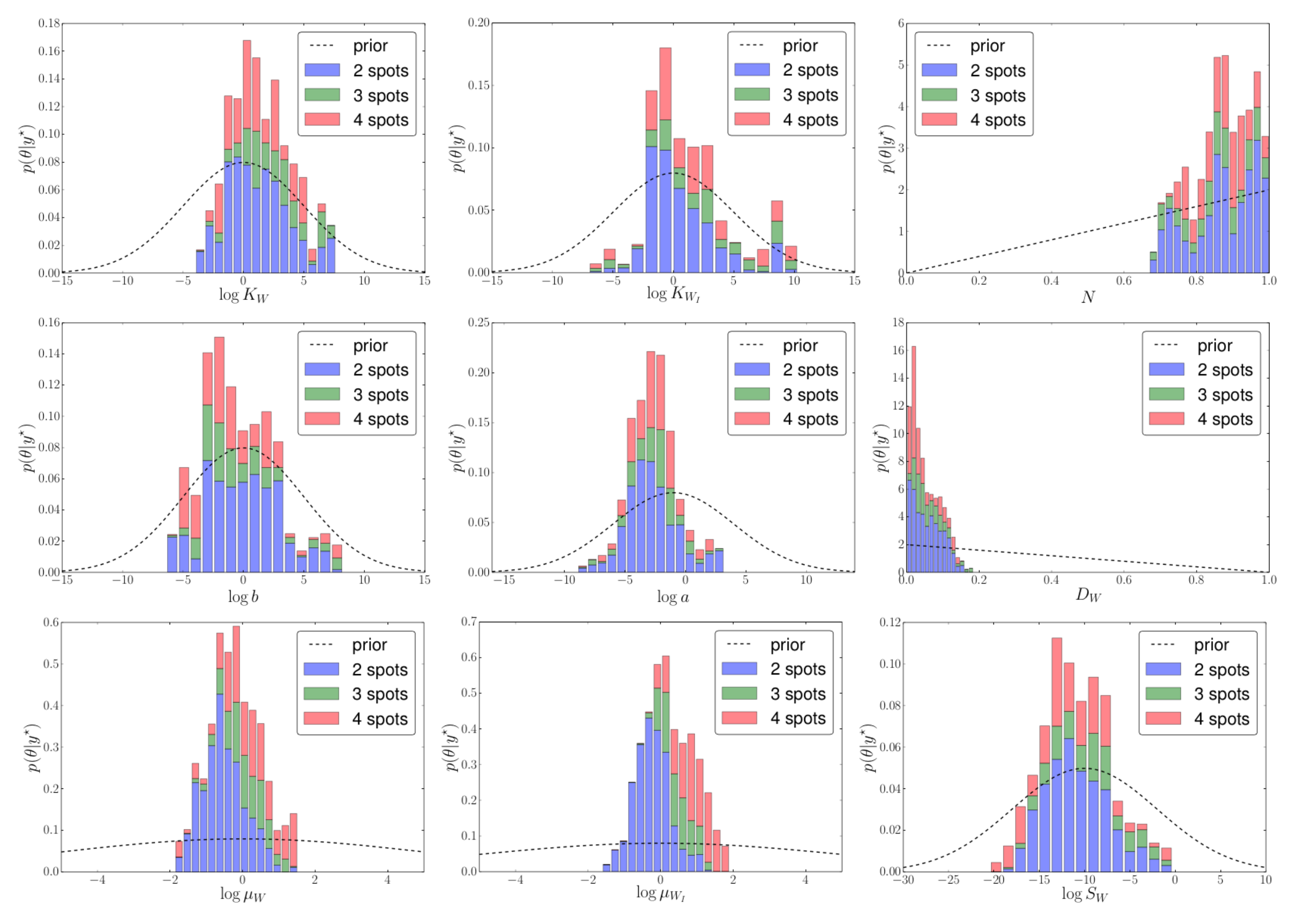}
\end{center}
\vspace{-0.2in}
\caption{\small{Marginal posterior parameter distributions.  Each figure shows the histogram for the marginal posterior distribution using colors {\em blue} ($\y_2 = 4$), {\em green} ($\y_2 = 3$), and {\em red} ($\y_2 = 2$) to differentiate spot numbers (associated with the simulator statistics run at their parameter setting). The prior $p(\thetav)$ is also shown as a dashed line.  Parameters $\mu_w$ and $\mu_{WI}$ have the most distinct spot-conditional distributions.}}
\label{fig:spots_post_marginal}
\end{figure}

Analysis of the posterior predictive distribution revealed distinct distributions when conditioned on $\y_2$, the number of spots.  The posterior distribution can therefore be viewed as a mixture of 3 spotting patterns, with $p(\y_2\,|\,\ystar ) = [ 0.505, 0.185, 0.310 ]$, where $\y_2 \in \{2,3,4\} $.  The marginal posterior predictive distributions are shown in Figure~\ref{fig:spots_ppd_pairwise} for pairs of statistics, and in Figure~\ref{fig:spots_ppd_marginal} for marginal distributions.  
To illustrate the role of the spotting patterns, by visual inspection of the posterior predictive distributions displayed in Figure \ref{fig:spots_ppd_pairwise}, we selected statistics labeled $A$ through $E$.  
  Parameters $\thetav$ corresponding to the modes $A$-$E$ were ran in both the 1D and 2D simulator producing images in Figure~\ref{fig:spots_at_fifty}, showing the desired shift away from Mock patterns.  Spot distributions were also found for $p(\thetav\,|\,\ystar)$, most distinctly for the Wnt and Wnt inhibitor decay rates ($\mu_W$ and $\mu_{W_I}$, respectively), which showed decreasing value for fewer spots, validating the original experimental results that blocking Wnt production by dnLEF-1 overexpression leads to qualitatively different spotting patterns.  The marginal posteriors are shown in Figure \ref{fig:spots_post_marginal}, along with the prior, for reference.   The strong relationship between $\mu_W$ and $\mu_{W_I}$ is shown in Figure \ref{fig:spots_muw_muwi}.  Subsamples from the posterior are overlaid with markers indicating the number of spots.   

\def \thisscale{0.18}

\def \thisscale{0.18}

\def \thisscale{0.18}

\def \thisscale{0.38}
 \begin{figure}[h]
 \begin{center}
 \includegraphics[scale=\thisscale]{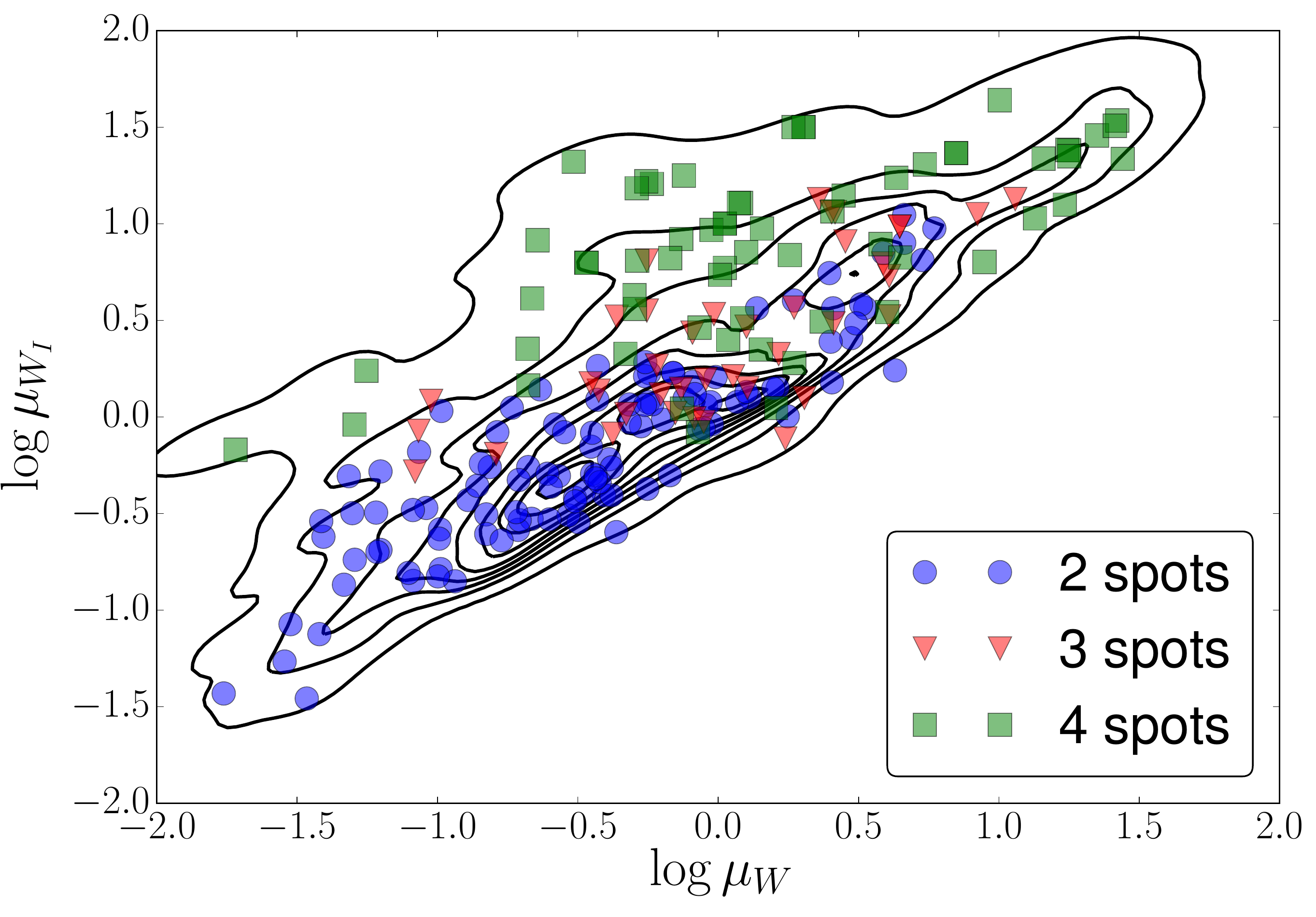}
 \end{center}
 \caption{{\small The posterior distribution of $\log \mu_{W}$ versus $\log \mu_{W_I}$.  Overlaid are subsamples from the posterior with colored symbols indicating the number of spots its setting produced, showing the strong relationship between these parameters and the number of spots.}}\label{fig:spots_muw_muwi}
 \end{figure}

This case study illustrates the usefulness of POPE for exploratory simulation analysis.  As a first attempt at studying this simulator from an ABC perspective, POPE revealed several regions of parameter settings that produce qualitatively different images from Mock.   Now experts can examine these various solutions to further develop the simulator or to increase the number of statistics.  For example, some of the parameter settings in the posterior seem to be similar to the prior, indicating they have little influence on the posterior.  If this does not match the intuition of the experts, the role these parameters have the simulator can be re-evaluated.  The $J=4$ statistics may also not be the most informative for the experts; based on our results learning the statistics (using computer vision techniques applied to the images) or modifying the current statistics may improve the ability of the experts to learn more about the spot formation process.

\section*{CONCLUSION}
There is considerable excitement in the machine learning community about optimizing objectives that are hard to evaluate, such as those defined by simulators. However, there is almost no work on analyzing such problems ``post optimization". We have found that this is exactly what scientists desire in order to study parameter dependencies and sensitivities and to compare different models in terms of their goodness of fit. We propose a post optimization posterior evaluation tool, POPE, by extending likelihood-free (ABC) MCMC samplers with one-sided kernels.  

Two case studies conducted in close collaboration with biologists show the usefulness of this new modeling framework.   For these studies we applied POPE in an optimization setting (stem-cell niche geometry) and a non-optimization setting (spotting patterns in cancer tissue), showing its usefulness for with {\em general}  constraint-based likelihoods.   
These preliminary results offer many avenues for future work.  Simulations with the stem-cell model could address whether giving cells some flexibility in the position at which they differentiate allows for more flexibility in the optimal geometry, perhaps allowing that geometry to also satisfy competing performance objectives.   Ongoing research with a modified version of the tumor metabolism  simulator will include non-constant nutrient levels and various therapeutic regimes, which will improve our understanding of cancer metabolism, and in turn  aid the development of new treatments or therapies.

\clearpage
\newpage
{
\bibliographystyle{plos2009}
\bibliography{optabc}
}
\clearpage
\section{POPE}\label{sec:appendixpope}

\begin{algorithm}[h] 
	\caption{Adaptive POPE}
	\label{algo:adaptivepope}
	\begin{algorithmic}[1]
  \Function{MCMC}{ $\thetav_0$, T, S, marginal, $\ystar$, $\yema$, $\gamma$, $\epsvec$, $\epsema$, $\epsmin$, $\delta$, $\epsquantile$}
    \State $\thetav \gets \thetav_0$
    \For{$t = 1 : T$}
      \State $\thetapv \sim q(\thetapv | \thetav )$
      \State $\ypone, \ldots, \ypS \overset{\simulator}{\sim} \pi( \y | \thetavp)$
      \If{marginal}
        \State $\yone, \ldots, \yS \overset{\simulator}{\sim} \pi( \y | \thetav)$ \Comment{Marginal samplers do not keep simulations.}
      \EndIf
      \State $\alpha \gets \lp 1, \frac{\pi(\thetavp)q(\thetap | \thetavp ) \pi(\thetavp | \ystar, \epsvec)}{ \pi(\thetav) q(\thetapv | \thetap )\pi(\thetav | \ystar, \epsvec)} \rp$
      \If{$\mathcal{U}(0,1) < \alpha$}
        \State $\thetav \gets \thetavp$
        \State $\yone, \ldots, \yS \gets \ypone, \ldots, \ypS$
      \EndIf
      \State $\ymu \gets \E\lb \yone, \ldots, \yS \rb$
      \State $\ystar, \yema \gets $ {\sc UpdateObjectives}$\lp \ystar, \ymu, \yema, \gamma \rp$ \Comment{See text, section \ref{sec:adapting}}
      \State $\epsvec, \epsema \gets $ {\sc UpdateEpsilons}$\lp \ystar, \ymu, \epsema, \epsmin, \delta, \epsquantile \rp $ \Comment{See text, section \ref{sec:adapting}}
      \State Collect $\thetav, \ymu, \epsvec, \ystar$\Comment{For posterior analysis.}
    \EndFor
     \State {\bf return} Collections $\thetav, \ymu, \epsvec, \ystar$
  \EndFunction
  \Function{UpdateObjectives}{ $\ystar, \y, \yema, \gamma$} \Comment{For $S>1$, $\y$ is the average.}
    \For{j = 1 : J}
      \State $\yemaj \gets (1-\gamma)\yemaj + \gamma \yj$ \Comment{Set $\gamma \gets 1$ for no update.}
      \State $\ystarj \gets \min\lp \ystarj, \yemaj \rp$ \Comment{Assume minimization.}
    \EndFor
    \State {\bf return} $\ystar, \yema$
  \EndFunction
  \Function{UpdateEpsilons}{ $\epsvec, \ystar, \y, \epsema, \epsmin, \delta, \epsquantile $}
    \For{j = 1 : J}
      \State $\Delta_j \gets (\ystarj - \yj)\textsc{Heavyside}\lp\ystarj - \yj\rp$ \Comment{Assume minimization.}
      \State $\epsemaj \gets (1-\delta)\yemaj + \delta \Delta_j$ \Comment{Set $\delta \gets 1$ for no update.}
      \State $\epsj \gets \max\lp \epsminj, \epsquantile \epsemaj \rp$ \Comment{Quantile $0 < \epsquantile < 1$ puts pressure on constraints. }
    \EndFor
    \State {\bf return} $\epsvec, \epsmin$
  \EndFunction
	\end{algorithmic}
\end{algorithm}

\clearpage 

\section{Spots}

\def \thisscale{0.33}
\begin{figure}[h]
  \vspace{-0.2in}
\begin{center}

  \includegraphics[width=\columnwidth]{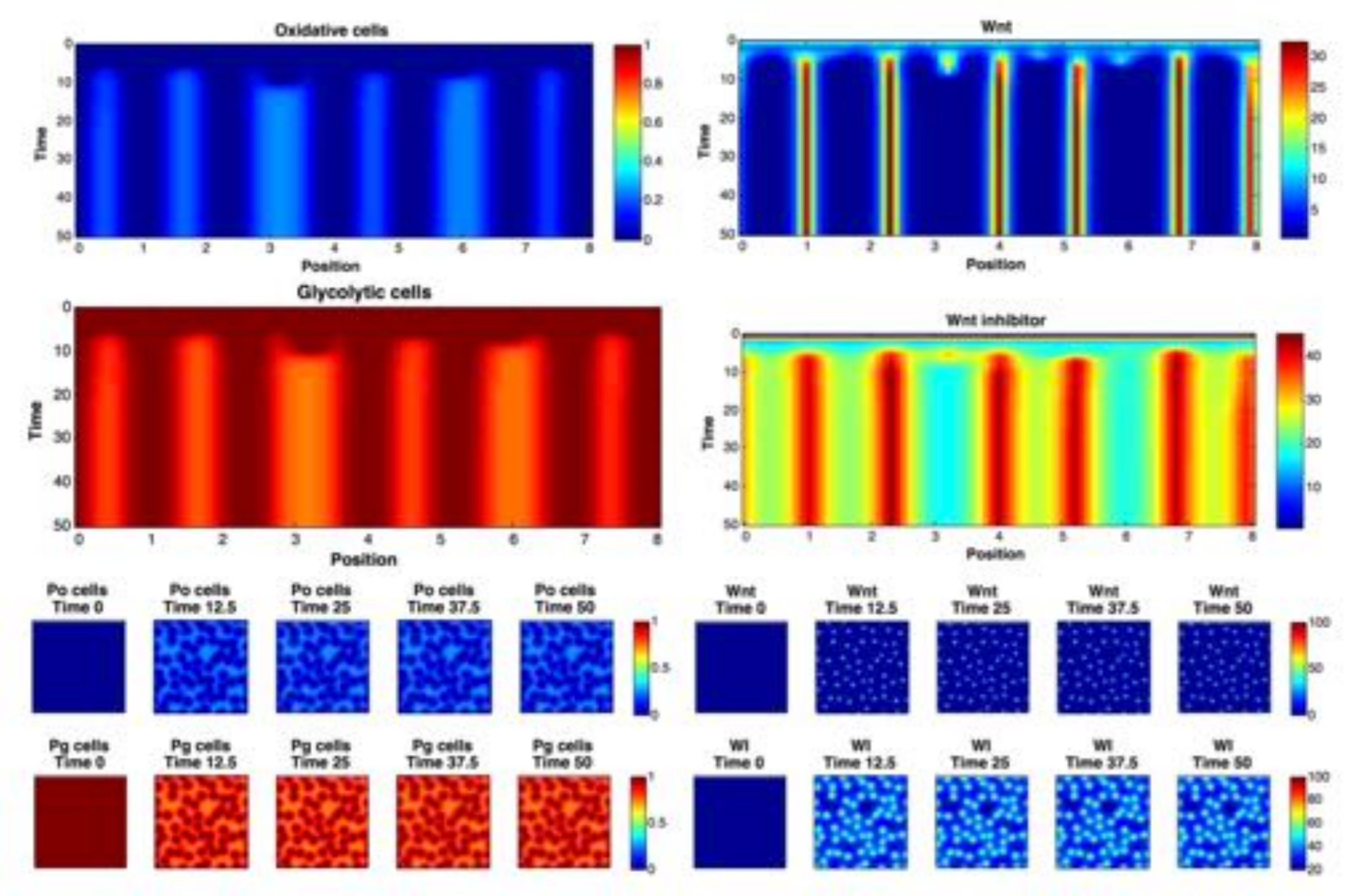}
\end{center} \vspace{-0.25in}
\caption{\small{Simulator outputs for the Mock setting of $\thetav$.  {\bf Upper plots:} 1D simulation results.  Images show the concentration of oxidative, glycolytic cells (left) and concentration of Wnt and Wnt inhibitor (right),  spatially and temporally.  {\bf Lower plots:} 2D simulator results.  Temporal slices of 2D spatial concentrations of oxidative (Po), glycolytic (Pg) cells (left) and Wnt and Wnt inhibitor (right).}}
\label{fig:spots_pp_mock_full}
\end{figure}

\begin{figure}[h]
\vspace{-0.1in}
\begin{center}
  \includegraphics[width=0.9\columnwidth]{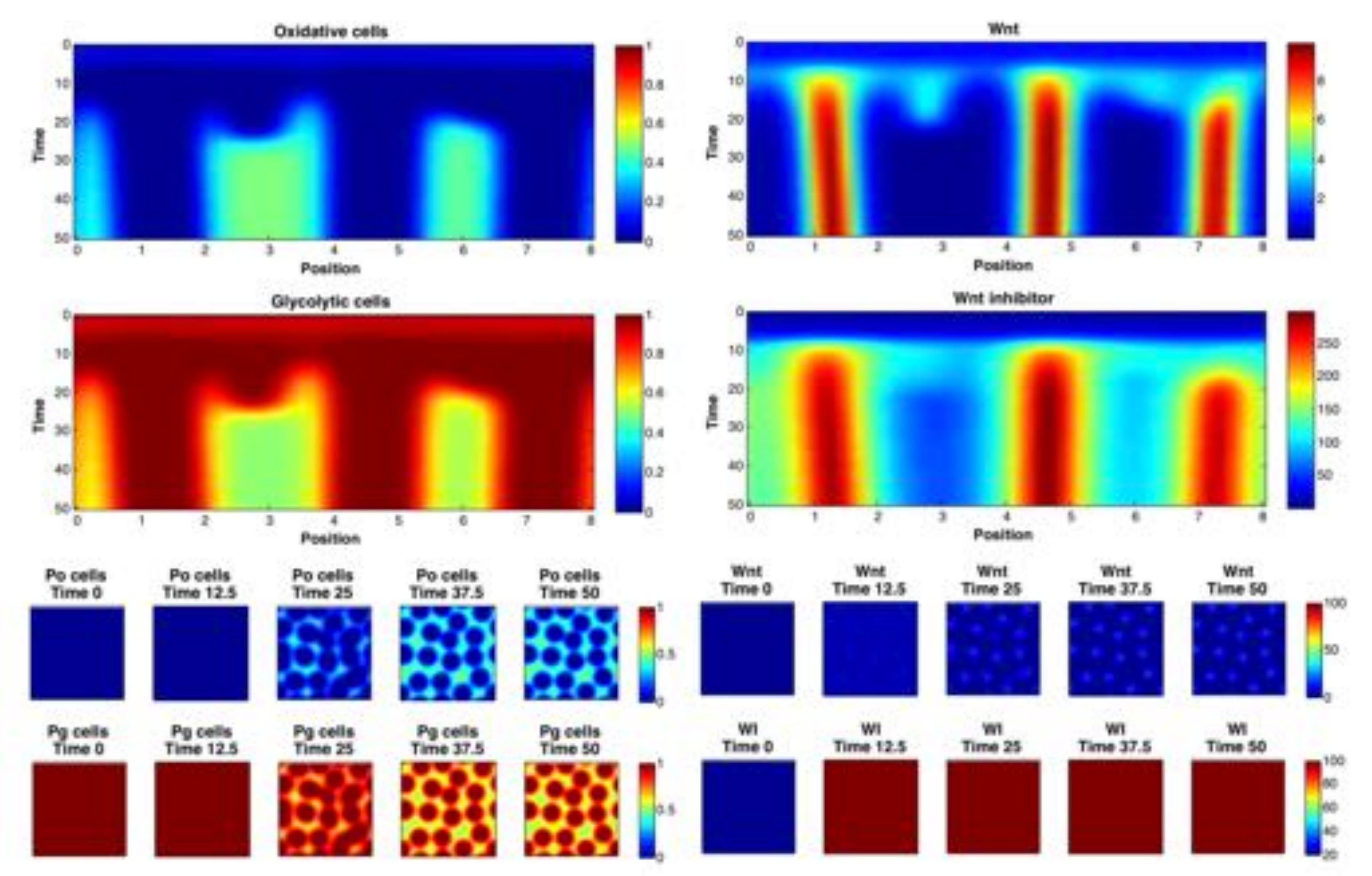}
\end{center}
\vspace{-0.25in}
\caption{\small{Simulator outputs for $\thetav$ setting $A$.}}
\label{fig:spots_pp_thetas_a_full}
\end{figure}

\begin{figure}[h]
\begin{center}
  \includegraphics[width=0.9\columnwidth]{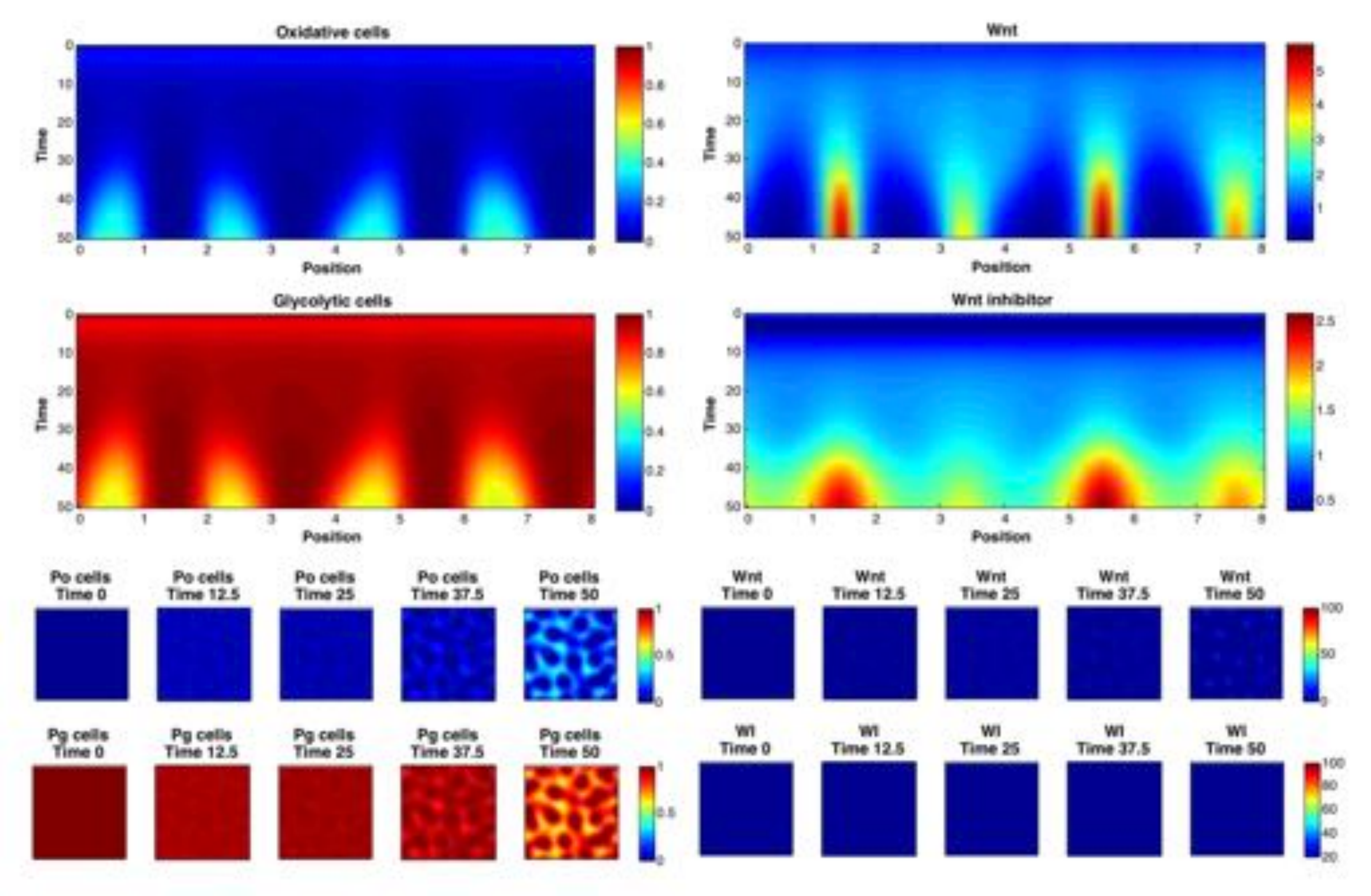}
\end{center}
\vspace{-0.25in}
\caption{\small{Simulator outputs for $\thetav$ setting $B$.}}
\label{fig:spots_pp_thetas_b_full}
\end{figure}

\begin{figure}[h]
\begin{center}
  \includegraphics[width=0.9\columnwidth]{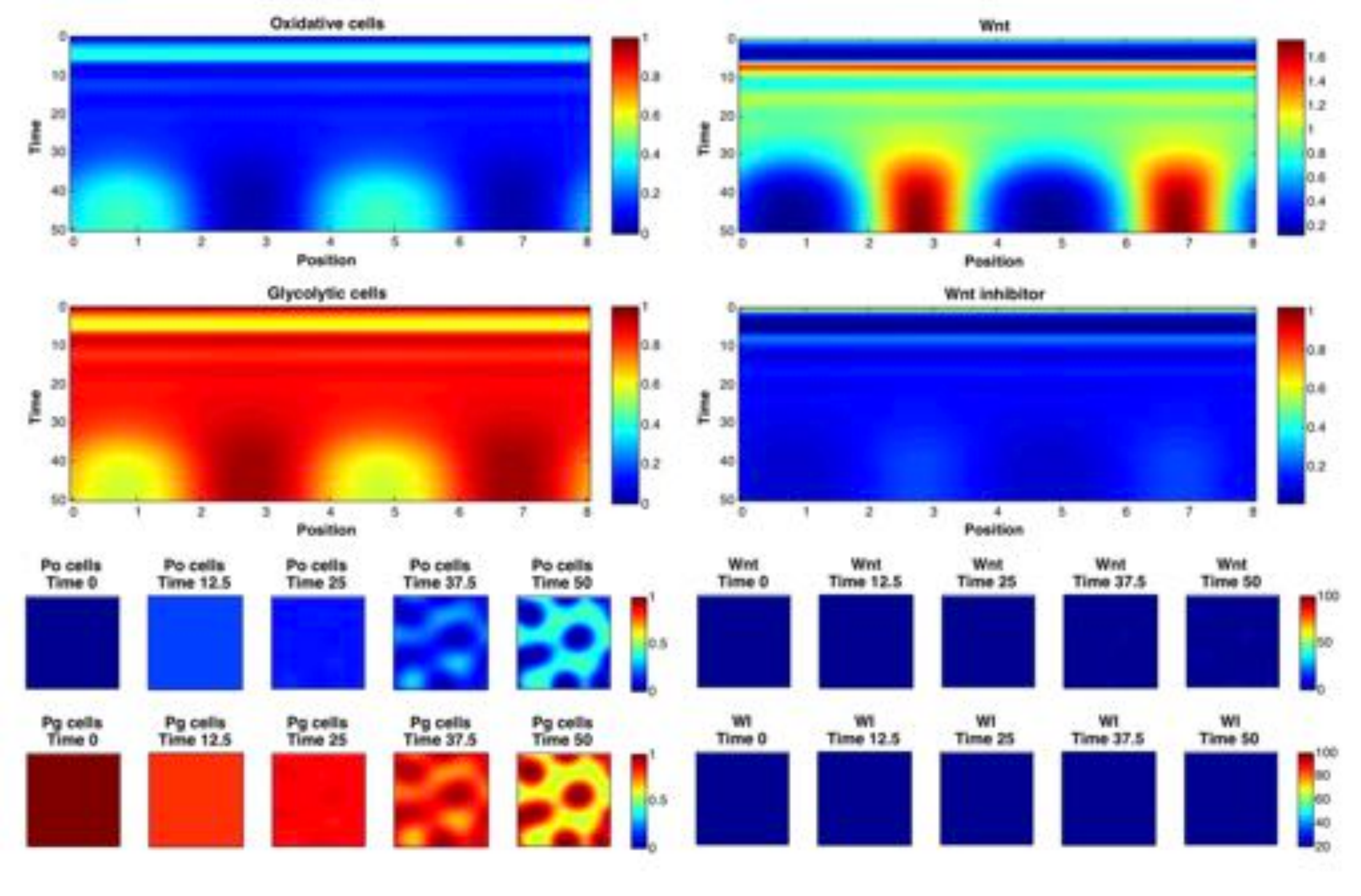}
\end{center}
\vspace{-0.25in}
\caption{\small{Simulator outputs for $\thetav$ setting $C$.}}
\label{fig:spots_pp_thetas_c_full}
\end{figure}

\begin{figure}[h]
\begin{center}
  \includegraphics[width=0.9\columnwidth]{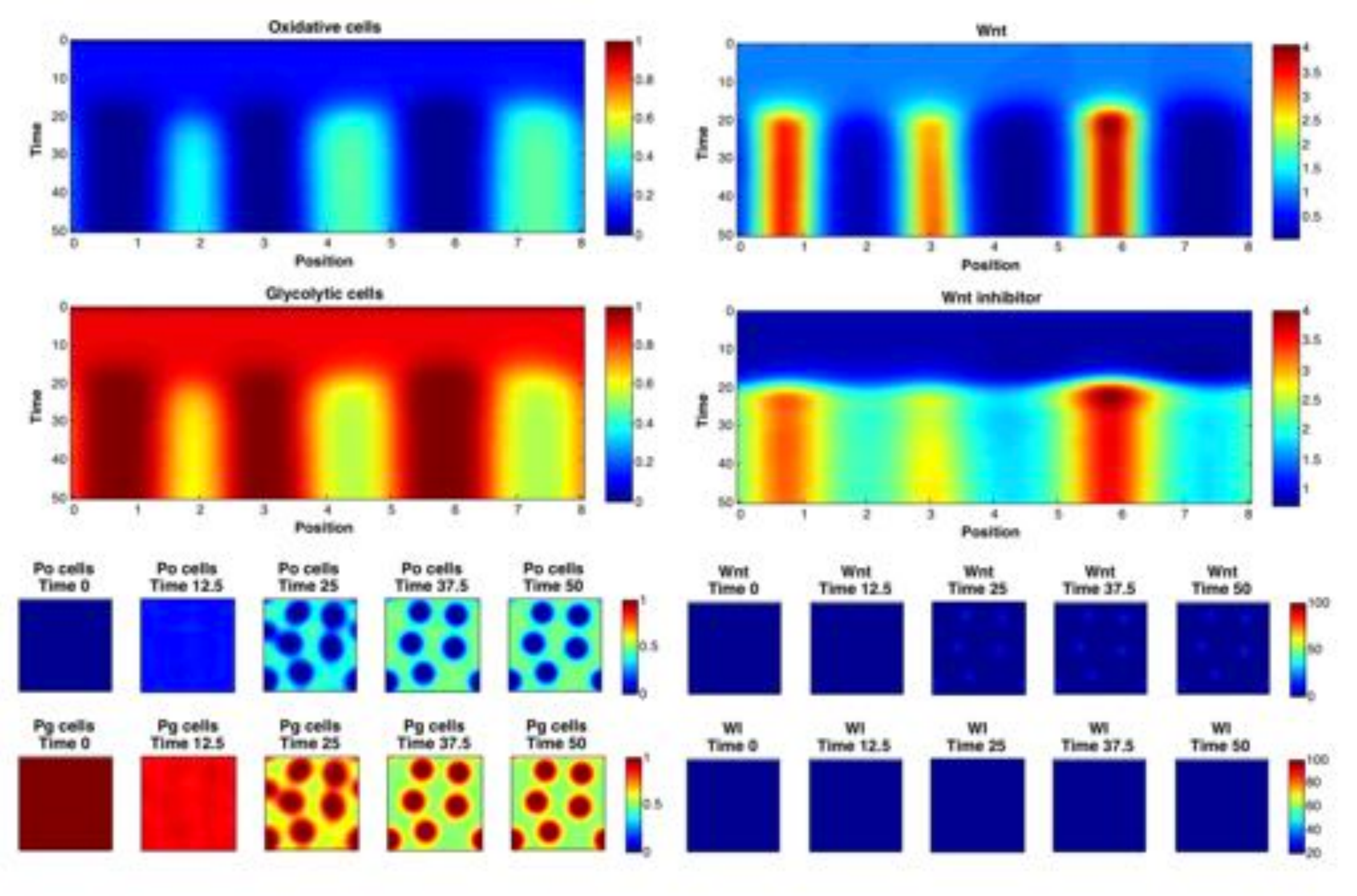}
\end{center}
\vspace{-0.25in}
\caption{\small{Simulator outputs for $\thetav$ setting $D$.}}
\label{fig:spots_pp_thetas_d_full}
\end{figure}

\begin{figure}[h]
\begin{center}
  \includegraphics[width=0.9\columnwidth]{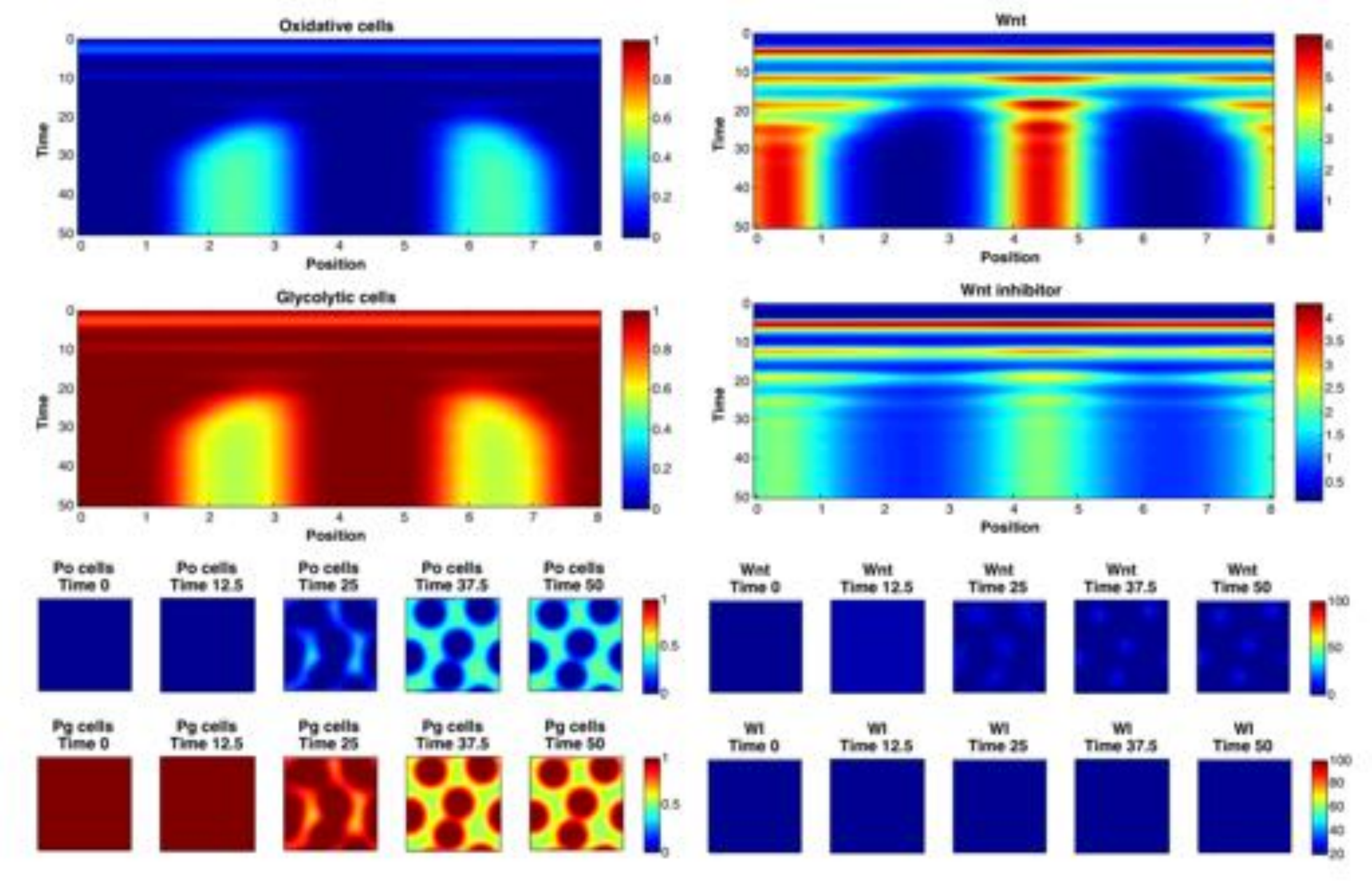}
\end{center}
\vspace{-0.25in}
\caption{\small{Simulator outputs for $\thetav$ setting $E$.}}
\label{fig:spots_pp_thetas_e_full}
\end{figure}

\end{document}